\documentclass[10pt,journal,compsoc]{IEEEtran}

\usepackage[table,dvipsnames]{xcolor}
\usepackage{tikz}  
\usepackage{times}
\usepackage{graphicx}
\usepackage{amsmath}
\usepackage{amssymb}
\usepackage{textcomp}

\usepackage{subfiles}
\usepackage{pgfplots} 
\usepackage{pgfplotstable} 
\pgfplotsset{compat=newest} 
\usetikzlibrary{plotmarks} 
\usetikzlibrary{colorbrewer} 
\usepackage{booktabs} 
\usepackage{etoolbox,siunitx} 

\robustify\bfseries
\robustify\itshape
\usepackage{paralist} 

\definecolor{olivegreen}{RGB}{0,170,0}
\definecolor{darkred}{RGB}{220,100,10}
\definecolor{tealblue}{RGB}{20,100,200}

\newcommand{\osvos}{Caelles2017}
\newcommand{\onavos}{Voigtlaender2017}
\newcommand{\ctn}{Jang2017}
\newcommand{\vpn}{Jampani2017}
\newcommand{\msk}{Perazzi2017}
\newcommand{\ofl}{Tsai2016}
\newcommand{\bvs}{NicolasMaerki2016}
\newcommand{\fcp}{Perazzi2015}
\newcommand{\jmp}{Fan2015}
\newcommand{\hvs}{Grundmann2010}
\newcommand{\sea}{Ramakanth2014}
\newcommand{\tsp}{Chang2013}

\newcommand{\arp}{Koh2017}
\newcommand{\fseg}{Jain2017}
\newcommand{\lmp}{Tokmakov2017}
\newcommand{\fst}{Papazoglou2013}

\newcommand{\msg}{Brox2010}

\newcommand{\nlc}{Faktor2014}

\newcommand{\ours}{OSVOS}
\newcommand{\oursnew}{OSVOS$^\textrm{S}$}

\newcommand{\longoursnew}{Semantic One-Shot Video Object Segmentation}

\newcommand{\nosvos}{OSVOS}
\newcommand{\nonavos}{OnAVOS}
\newcommand{\nmsk}{MSK}
\newcommand{\nctn}{CTN}
\newcommand{\nvpn}{VPN}
\newcommand{\nofl}{OFL}
\newcommand{\nbvs}{BVS}
\newcommand{\nfcp}{FCP}

\newcommand{\narp}{ARP}
\newcommand{\nfseg}{FSEG}
\newcommand{\nlmp}{LMP}
\newcommand{\nfst}{FST}

\newcommand{\nmsg}{MSG}

\newcommand{\nnlc}{NLC}

\newcommand{\ncobspub}{COB|SP}

\newcommand{\J}{\mathcal{J}}
\newcommand{\F}{\mathcal{F}}
\newcommand{\T}{\mathcal{T}}

\definecolor{rowblue}{RGB}{220,230,240}

\newcommand{\ie}{i.e.\ } 
\newcommand{\eg}{e.g.\ } 
\newcommand{\etal}{et al.\ } 

\usepackage[pagebackref=true,breaklinks=true,letterpaper=true,colorlinks,bookmarks=false]{hyperref}

\ifCLASSOPTIONcompsoc
  \usepackage[nocompress]{cite}
\else
  \usepackage{cite}
\fi

\ifCLASSINFOpdf

\else

\fi

\begin{document}

\title{Video Object Segmentation\\Without Temporal Information}

\author{K.-K. Maninis*, S. Caelles*, Y. Chen,\\J. Pont-Tuset, L. Leal-Taix\'e, D. Cremers, and L. Van Gool
\IEEEcompsocitemizethanks{\IEEEcompsocthanksitem K.-K. Maninis, S. Caelles, Y. Chen, J. Pont-Tuset, and L. Van Gool are with the ETHZ, Z\"urich. First two authors contributed equally.\protect
\IEEEcompsocthanksitem L. Leal-Taix\'e and D. Cremers are with the TUM, M\"unchen.\protect
\IEEEcompsocthanksitem Contacts in http://www.vision.ee.ethz.ch/\~{}cvlsegmentation/}
}

\markboth{}
{}
\IEEEtitleabstractindextext{%
\begin{abstract}
Video Object Segmentation, and video processing in general, has been historically dominated by methods that rely on the temporal consistency and redundancy in consecutive video frames.
When the temporal smoothness is suddenly broken, such as when an object is occluded, or some frames are missing in a sequence, the result of these methods can deteriorate significantly. This paper explores the orthogonal approach of processing each frame independently, \ie disregarding the temporal information. In particular, it tackles the task of semi-supervised video object segmentation: the separation of an object from the background in a video, given its mask in the first frame.
We present \longoursnew{} (\oursnew{}), based on a fully-convolutional neural network architecture that is able to successively transfer generic semantic information, learned on ImageNet, to the task of foreground segmentation, and finally to learning the appearance of a single annotated object of the test sequence (hence one shot). We show that instance-level semantic information, when combined effectively, can dramatically improve the results of our previous method, \ours{}.
We perform experiments on two recent single-object video segmentation databases, which show that \oursnew{} is both the fastest and most accurate method in the state of the art. Experiments on multi-object video segmentation show that \oursnew{} obtains competitive results.
\end{abstract}

}

\maketitle

\IEEEdisplaynontitleabstractindextext

\IEEEpeerreviewmaketitle

\IEEEraisesectionheading{\section{Introduction}\label{sec:intro}}
\IEEEPARstart{A}{}video is a temporal sequence of static images that give the impression of continuous motion
when played consecutively and rapidly.
The illusion of motion pictures is due to the persistence of human vision~\cite{Hopwood1899,Wang2001,Tekalp2015}: the fact that it cannot perceive very high frequency changes~\cite{Wang2001} because of the temporal integration of incoming light into the retina~\cite{Tekalp2015}.
This property has been exploited since the appearance of the phenakistoscope~\cite{Stampfer1833} or the zoetrope~\cite{Hopwood1899}, which displayed a sequence of drawings creating the illusion of continuous movement.

In order to achieve the high frequency to produce the video illusion, consecutive images vary very smoothly and slowly: the information in a video is very redundant and neighboring frames carry very similar information.
In video coding, for instance, this is the key idea behind video compression algorithms such as motion-compensated coding~\cite{Tekalp2015}, where instead of storing each frame independently, one picks a certain image and only codes the modifications to be done to it to generate the next frame.

Video processing in general, and video segmentation in particular, is also dominated by this idea,
where {\it motion estimation} has emerged as a key ingredient for some of the state-of-the-art video segmentation 
algorithms~\cite{JaKi17, Perazzi2017,Tsai2016,Ramakanth2014,Grundmann2010}. Exploiting it is not a trivial task 
however, as one has to compute temporal matches in the form of optical flow or dense 
trajectories~\cite{Brox2010}, which can be an even harder problem to solve.

On the other hand, processing each frame independently would allow us to easily parallelize the computation, and to not be affected by sequence interruptions, to process the frames at any desire rate, etc.
This paper explores how to segment objects in videos when processing each frame independently, that is, by ignoring the temporal information and redundancy.
In other words, we cast video object segmentation as a per-frame segmentation problem given
the \textit{model} of the object from one (or various) manually-segmented frames.

This stands in contrast to the dominant approach where temporal consistency plays the central role, assuming that objects do not change too much between one frame and the next. Such methods adapt their single-frame models smoothly throughout the video, looking for targets whose shape and appearance vary \textit{gradually} in consecutive frames, but fail when those constraints do not apply, unable to recover from relatively common situations such as occlusions and abrupt motion.

We argue that temporal consistency was needed in the past, as one had to overcome major drawbacks of the then inaccurate shape or appearance models.
On the other hand, in this paper deep learning will be shown to provide a sufficiently accurate model of the target object to produce very accurate results even when processing each frame independently.
This has some natural advantages: \oursnew{} is able to segment objects throughout occlusions, it is not limited to certain ranges of motion, it does not need to process frames sequentially, and errors are not temporally propagated.
In practice, this allows \oursnew{} to handle \eg interlaced videos of surveillance scenarios, where cameras can go blind for a while before coming back on again.

\begin{figure*}
      \setlength{\fboxsep}{0pt}
      \resizebox{\textwidth}{!}{%
        \fbox{\includegraphics{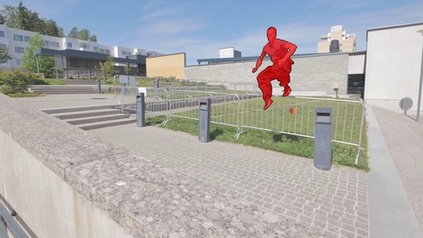}}\hspace{4mm}
       \fbox{\includegraphics{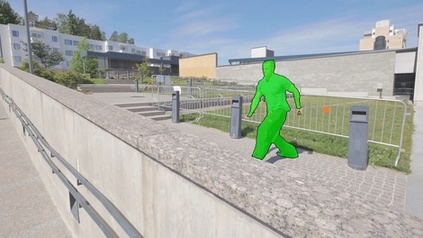}}\hspace{4mm}
       \fbox{\includegraphics{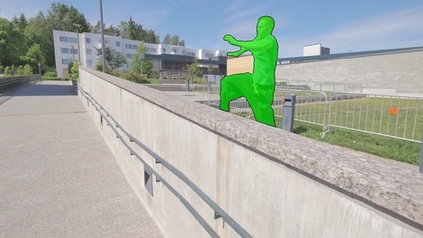}}\hspace{4mm}
       \fbox{\includegraphics{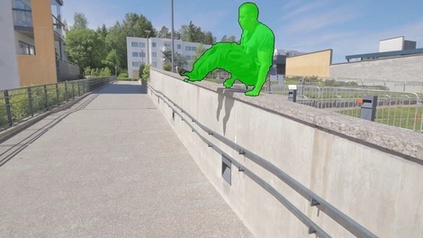}}\hspace{4mm}
       \fbox{\includegraphics{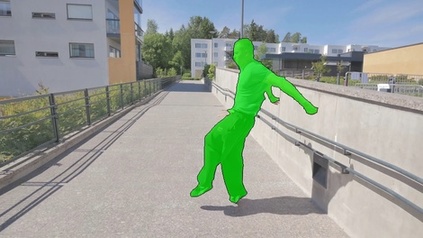}}
       }\\[2mm]
       \resizebox{\textwidth}{!}{%
       \fbox{\includegraphics{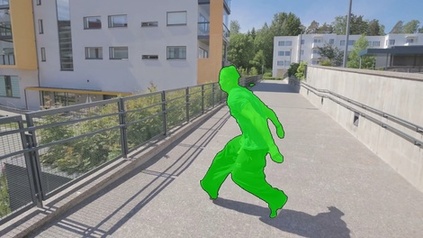}}\hspace{4mm}
       \fbox{\includegraphics{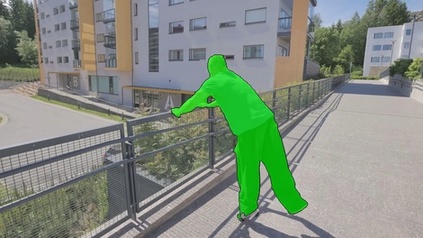}}\hspace{4mm}
       \fbox{\includegraphics{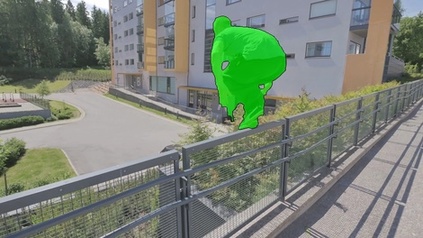}}\hspace{4mm}
       \fbox{\includegraphics{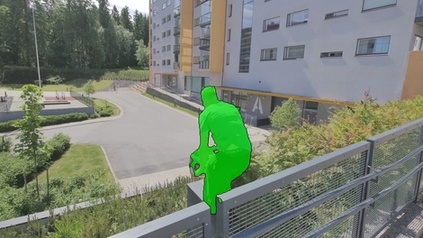}}\hspace{4mm}
       \fbox{\includegraphics{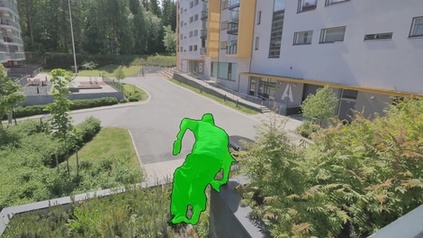}}
      }
       \caption{\textbf{Example result of our technique:} The segmentation of the first frame (red) is used to learn the model of the specific object to track, which is segmented in the rest of the frames independently (green). One every 10 frames shown of 90 in total.}
      \label{fig:example}
\end{figure*}

Given the first frame, we create an \textit{appearance model} of the object of interest and then look for the pixels that better match this model in the rest of the frames.
To do so, we will make use of Convolutional Neural Networks (CNNs), which are revolutionizing many fields of computer vision. For instance, they have dramatically boosted the performance for problems like image classification~\cite{Krizhevsky2012,SiZi15,He+16} and object detection~\cite{Girshick2014,Gir15,Liu+16}. Image segmentation has also been taken over by CNNs recently~\cite{Man+17,Kokkinos2016,XiTu15,BST15b,BST16},
with deep architectures pre-trained on the weakly related task of image classification on ImageNet~\cite{Russakovsky2015}. One of the major downsides of deep network approaches, however, is their hunger for training data. Yet, with various pre-trained network architectures one may ask how much training data do we really need for the specific problem at hand? This paper investigates segmenting an object along an entire video, when we only have one single labeled training example, e.g. the first frame.

Figure~\ref{fig:example} shows an example result of \oursnew{}, where the input is the segmentation of the first frame
(in red), and the output is the mask of the object in the 90 frames of the sequence (in green).

The first contribution of the paper is to adapt the CNN to a particular object instance
given a single annotated image.
To do so, we gradually adapt a CNN pre-trained on image recognition~\cite{Russakovsky2015} to video object segmentation. This is achieved by training it on a set of videos with manually segmented objects. Finally, it is fine-tuned \textit{at test time} on a specific object that is manually segmented in a single frame.
Figure~\ref{fig:overview} shows the overview of the method.
Our proposal tallies with the observation that 
leveraging these different levels of information
to perform object segmentation would stand to reason: from generic information of a large
amount of categories, passing through the knowledge of the \textit{usual} shapes of objects in videos,
down to the specific properties of a particular object we are interested in segmenting.

Our second contribution is to extend the model of the object with explicit semantic information.
In the example of Figure~\ref{fig:example}, for instance, we would like to leverage the fact that we are segmenting an object of the category \textit{person} 
and that there is a \textit{single instance} of it.

In particular, we will use an \textit{instance-aware semantic segmentation algorithm}~\cite{dai2016instance, Li+17, He+17} to extract a list of proposal of object masks in each frame, along with their categories.
Given the first annotated frame, we will \textit{infer} the categories of the objects of interest by finding the best-overlapping masks.
We refer to this step as ``semantic selection.''

Our method uses the extracted semantic information from the first frame to segment the rest of the video. 
It enforces the resulting masks to align well with the same categories selected in the first frame.
If we were segmenting a person on a motorbike, then this information should be kept throughout the video.
In particular, we find instances extracted from the semantic instance segmentation algorithm that best match the model of the object, and we effectively combine them with the appearance model of the object, using a conditional classifier.
We call this step ``semantic propagation.''

Our third contribution is that \oursnew{} can work at various points of the trade-off between
speed and accuracy.
In this sense, given one annotated frame, the user can choose the level of fine-tuning performed on it,
giving them the freedom between a faster method or more accurate results.
Experimentally, we show that \oursnew{} can run at 300 miliseconds per frame and 75.1\% accuracy, and up to 86.5\% when processing each frame in 4.5 seconds, for an image of 480$\times$854 pixels.

Technically, we adopt the architecture of Fully Convolutional Networks (FCN)~\cite{farabet2013learning, LSD15}, suitable for dense predictions. FCNs have recently become popular due to their performance both in terms of accuracy and computational efficiency~\cite{LSD15,Dai2016a,Dai2016}.
Arguably, the Achilles' heel of FCNs when it comes to segmentation is the coarse scale of the deeper layers, which leads to inaccurately localized predictions.
To overcome this, a large variety of works from different fields use skip connections of larger feature maps~\cite{LSD15,Har+15,XiTu15,Man+16}, 
or learnable filters to improve upscaling~\cite{NHH15,Yan+16}.

We perform experiments on two video object segmentation datasets (DAVIS 2016~\cite{Perazzi2016} and Youtube-Objects~\cite{Prest2012,Jain2014}) and show that \oursnew{} significantly improves the state of the art in them, both in terms of accuracy and speed. We perform additional experiments for multi-object video segmentation on DAVIS 2017~\cite{Pont-Tuset2017}, where we obtain competitive results by directly applying our method without adaptation to the new problem.

All resources of this paper, including training and testing code, pre-computed results, and pre-trained models will be made publicly available.

\section{Related Work}
\label{sec:related}

\paragraph*{\textbf{Semi-supervised Video Object Segmentation}}
Most of the current literature on semi-supervised video object segmentation enforces 
temporal consistency in video sequences to propagate the initial mask into the following frames.
The most recent works heavily rely on optical flow, and make use of CNNs to learn to refine the mask of the object at frame $n$ to frame $n+1$~\cite{\msk, \ctn} or combine the training of a CNN with ideas of bilateral filtering between consecutive frames~\cite{\vpn}. Also, \cite{\onavos} follows up with the idea introduced in OSVOS and uses the result on the the predicted frames on the whole sequence to further train the network at test time.
Previously, and in order to reduce the computational complexity, some works make use of superpixels~\cite{\tsp,\hvs}, patches~\cite{\sea,\jmp}, object proposals~\cite{\fcp}, or the bilateral space~\cite{\bvs}.
After that, an optimization using one of the previous aggregations of pixels is usually performed; which can consider the full video sequence~\cite{\fcp,\bvs}, a subset of frames~\cite{\hvs}, or only the results in frame $n$ to obtain the mask in $n\!+\!1$~\cite{\sea,\tsp,\jmp}.
As part of their pipeline, some of the methods include the computation of optical flow~\cite{\hvs,\sea,\msk}, or/and Conditional Random Fields (CRFs)~\cite{\msk} which can considerably reduce their speed.
Different from those approaches, \oursnew{} is a simpler pipeline which segments each frame independently, and produces more accurate results, while also being significantly faster.

\paragraph*{\textbf{FCNs for Segmentation}} 
Segmentation research has closely followed the innovative ideas of CNNs in the last few years. The advances observed in image recognition~\cite{Krizhevsky2012, SiZi15, He+16} have been beneficial to segmentation in many forms (semantic~\cite{LSD15, NHH15}, instance-level~\cite{Gir15,Pinheiro2016,Dai2016a}, biomedical~\cite{RFB15}, generic~\cite{Man+17}, etc.).
Many of the current best performing methods are based on a deep CNN architecture, usually pre-trained on ImageNet~\cite{Russakovsky2015}, trainable end-to-end. The idea of dense predictions with CNNs was pioneered by~\cite{farabet2013learning} and formulated by~\cite{LSD15} in the form of Fully Convolutional Networks (FCNs) for semantic segmentation. The authors noticed that by changing the last fully connected layers to $1\times1$ convolutions it is possible to train on images of arbitrary size, by predicting correspondingly-sized outputs.
Their approach boosts efficiency over patch-based approaches where one needs to perform redundant computations in overlapping patches. More importantly, by removing the parameter-intensive fully connected layers, the number of trainable parameters drops significantly, facilitating training with relatively fewer labeled data.

In most CNN architectures~\cite{Krizhevsky2012, SiZi15, He+16}, activations of the intermediate layers gradually decrease in size,
because of spatial pooling operations or convolutions with a stride. Making dense predictions from downsampled activations results in coarsely localized outputs~\cite{LSD15}. Deconvolutional layers that learn how to upsample are used in~\cite{NHH15,Yan+16} to recover accurately localized predictions.
In~\cite{Pinheiro2016}, activations from shallow layers are gradually injected into the prediction to favor localization. However, these architectures come with many more trainable parameters and their use is limited to cases with sufficient data.

Following the ideas of FCNs, Xie and Tu~\cite{XiTu17} separately supervised the intermediate layers of a deep network for contour detection. The duality between multiscale contours and hierarchical segmentation~\cite{Arb+11,Pont-Tuset2016} was further studied by Maninis \etal~\cite{Maninis2016} by bringing CNNs to the field of generic image segmentation. In this work we explore how to train an FCN for accurately localized dense prediction based on very limited annotation: a single segmented frame.

\paragraph*{\textbf{Semantic Instance Segmentation}} 
Semantic instance segmentation is a relatively new computer vision task which has recently gained increasing attention. In contrast to semantic segmentation or object detection, the goal of instance segmentation is to provide a segmentation mask for each individual instance. The task was first introduced in~\cite{hariharan2014simultaneous}, where they extract both region and foreground features using the R-CNN~\cite{Girshick2014} framework and region proposals. Then, the features are concatenated and classified by an SVM. Several works~\cite{dai2016instance,dai2015convolutional,zagoruyko2016multipath} following that path have been proposed in recent years. There also exist some approaches based on iteration~\cite{li2016iterative}, and recurrent neural networks~\cite{romera2016recurrent}. The recent best-performing methods use fully convolutional position sensitive architectures~\cite{Li+17}, or a modified Faster-RCNN~\cite{Ren+15} pipeline, extended to instance segmentation~\cite{He+17}. In contrast to such class-sensitive methods, in which unseen classes are treated as background, our method is class agnostic, and is able to segment generic objects, given only one annotated example.

\paragraph*{\textbf{Using Semantic Information to Aid Other Computer Vision Tasks}}
Semantic information is a very relevant cue for the human vision system, and some computer vision algorithms leverage it to aid various tasks. \cite{hane2013joint} improves reconstruction quality by jointly reasoning about class segmentation and 3D reconstruction. Using a similar philosophy, \cite{liu2010single} estimates the depth of each pixel in a scene from a single monocular image guided by semantic segmentation, and improves the results significantly.
To the best of our knowledge, we are the first ones to apply instance semantic information to the task of object segmentation in videos.

\paragraph*{\textbf{Conditional Models}}
Conditional models prove to be a very powerful tool when the feature statistics are complex.
In this way, prior knowledge can be introduced by incorporating a dependency to it.
\cite{dantone2012real} builds a conditional random forest to estimate face landmarks whose
classifier is dependent on the pose of head.
Similarly, \cite{sun2012conditional} proposes to estimate human pose dependent on torso orientation, or human height, which can be a useful cue for the task of pose estimation.
The same also applies to boundary detection, \cite{uijlings2015situational} proposes to train a series of conditional boundary detectors, and the detectors are weighted differently during test based on the global context of the test image.
In this work, we argue that the feature distribution of foreground and background pixels are essentially different, and so a monolithic classifier for the whole image is bound to be suboptimal.
Thus, we utilize the conditional classifier to better model the different distributions.

\section{One-Shot Video Object Segmentation (\ours)}
\label{sec:osvos}
This section describes our algorithm to gradually fine-tune the CNN in order to build a strong appearance model for video object segmentation given the first annotated frame. This was presented in our conference contribution~\cite{Caelles2017}. We will refer to the method as \ours{}, to differentiate it from \oursnew{} (Section~\ref{sec:semantic}), in which we use semantic instance segmentation as further guiding signal.

Let us assume that one would like to segment an object in a video, for which the only available piece of 
information is its foreground/background segmentation in one frame.
Intuitively, one could analyze the entity, create a \textit{model},
and search for it in the rest of the frames.
For humans, this very limited amount of information is more than enough, and changes in appearance, shape, 
occlusions, etc.\ do not pose a significant challenge, because we leverage strong priors: first ``It is an object,'' and then ``It is \textit{this particular} object.''
Our method is inspired by this gradual refinement.

\begin{figure*}
\centering
\vspace{-2mm}
\includegraphics[width=0.8\textwidth]{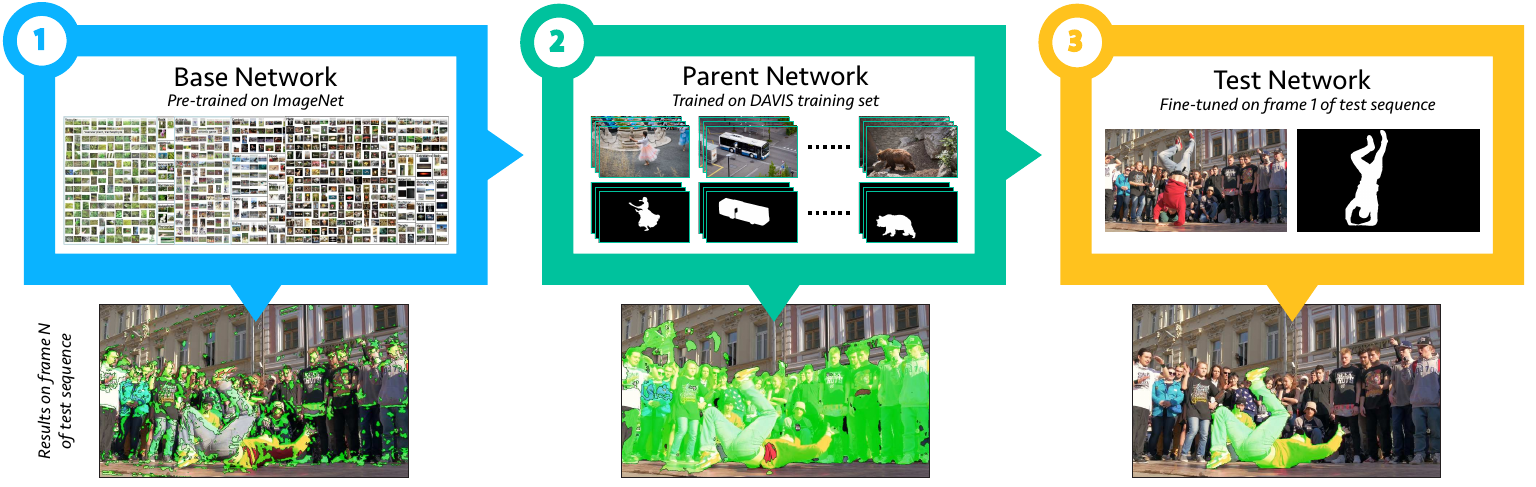}\\[1mm]
\caption{\textbf{Overview of \ours{}}: (1) We start with a pre-trained base CNN for image labeling on ImageNet; its results in terms of segmentation, although conform with some image features, are not useful.
(2) We then train a \textit{parent network} on the training set of DAVIS 2016; the segmentation results improve but are not focused on an specific object yet. (3) By fine-tuning on a segmentation example for the specific target object in a single frame, the network rapidly focuses on that target.}
\label{fig:overview}
\vspace{-2mm}
\end{figure*}

We train a Fully Convolutional Neural Network (FCN) for the binary classification task of separating the foreground object from the background.
We use two successive training steps: First we train on a large variety of objects, offline, to construct a 
model that is able to discriminate the general notion of a foreground object, \ie, ``It is an object.''
Then, at test time, we fine-tune the network for a small number of iterations on the particular
instance that we aim to segment, \ie, ``It is \textit{this particular} object.''
The overview of our method is illustrated in Figure~\ref{fig:overview}.

\subsection{End-to-end trainable foreground FCN}
Ideally, we would like our CNN architecture to satisfy the following criteria:
(i) Accurately localized segmentation output, as discussed in Section~\ref{sec:related},
(ii) relatively small number of parameters to train from a limited amount of annotated data, and
(iii) relatively fast testing times.

\begin{figure}
\includegraphics[width=\linewidth]{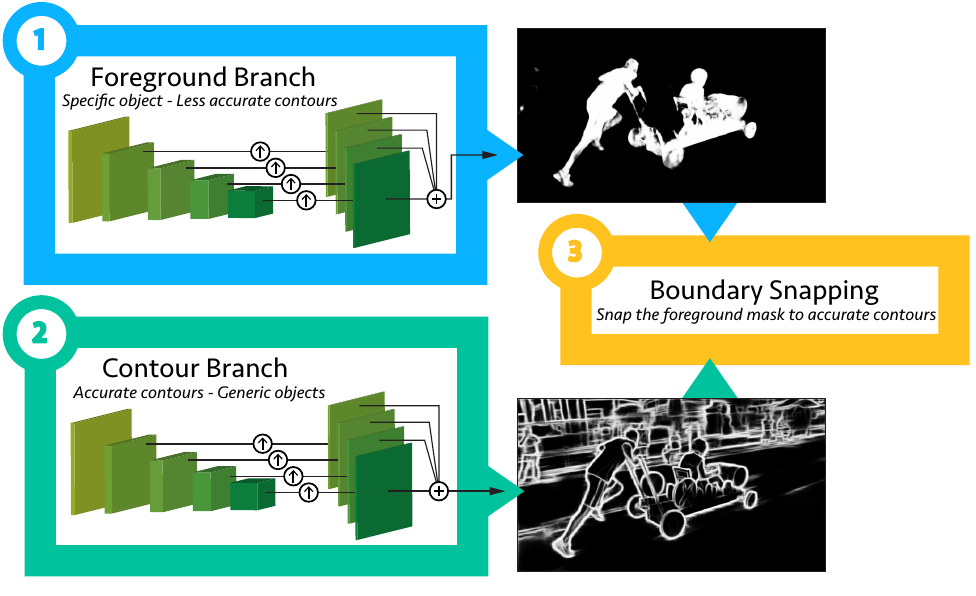}
\caption{\textbf{Two-stream FCN architecture}: The main foreground branch (1) is complemented by a contour branch (2) which improves the localization of the boundaries (3).}
\label{fig:two-stream}
\vspace{-2mm}
\end{figure}

We draw inspiration from the CNN architecture of~\cite{Man+16}, originally used for biomedical
image segmentation.
It is based on the VGG~\cite{SiZi15} network, modified for accurately localized dense prediction (Point i).
The fully-connected layers needed for classification are removed (Point ii), and efficient image-to-image 
inference is performed (Point iii).
The VGG architecture consists of groups of convolutional plus Rectified Linear Units (ReLU)~\cite{Nair2010}
layers grouped into 5 stages.
Between the stages, pooling operations downscale the feature maps as we go deeper into the network.
We connect convolutional layers to form separate skip paths from the last layer of each stage
(before pooling).
Upscaling operations take place wherever necessary, and feature maps from the separate paths are
concatenated to construct a volume with information from different levels of detail.
We linearly fuse the feature maps to a single output which has the same dimensions as the image,
and we assign a loss function to it.
The proposed architecture is shown in Figure~\ref{fig:two-stream} (1), foreground branch.

The pixel-wise cross-entropy loss for binary classification (we keep the notation of Xie and Tu~\cite{XiTu15}) is in this case defined as:

{\footnotesize
\begin{equation}
\begin{split}
\mathcal{L}\left(\mathbf{W}\right)\!&=\!-\!\!\sum_{j}\!{y_j\!\log{\!P\left(y_j\!\!=\!\!1 |X;\!\mathbf{W}\right)}\!+\!\left( 1\!\!-\!\!y_j\right)\!\log{\left(1\!\!-\!\!P\left(y_j\!\!=\!\!1 |X;\!\mathbf{W}\right)\!\right)}}\\
&=\!-\!\!\!\sum_{j \in Y_+}\!\!{\!\log{\!P\left(y_j\!\!=\!\!1 |X;\!\mathbf{W}\right)}}-\!\!\sum_{j	\in Y_-}\!\!{\!\log{\!P\left(y_j\!\!=\!\!0 |X;\mathbf{W}\right)}} \label{eq:cross_entropy}
\end{split}\nonumber
\end{equation}}\\[-4mm]
where $\mathbf{W}$ are the standard trainable parameters of a CNN, $X$ is the input image, $y_j \in \{0,1\}$, $j=1,\dots,|X|$ is the pixel-wise binary label of $X$, and $Y_+$ and $Y_-$ are the positive and negative labeled pixels. $P(\cdot)$ is obtained by applying a sigmoid to the activation of the final layer.

In order to handle the imbalance between the two binary classes, Xie and Tu~\cite{XiTu15} proposed a modified version of the cost function, originally used for contour detection (we drop $\mathbf{W}$ for the sake of readability):

{\footnotesize
\begin{align}
\mathcal{L}_{mod}\!=\!-\beta\!\!\sum_{j \in Y_+}\!\!{\!\log{\!P\left(y_j\!\!=\!\!1 |X\right)}}-(1\!-\!\beta)\!\!\sum_{j\in Y_-}\!\!{\!\log{\!P\left(y_j\!\!=\!\!0 |X\right)}} \label{eq:hed_cost}
\end{align}}%
where $\beta=|Y_-|/|Y|$. Equation~\ref{eq:hed_cost} allows training for imbalanced binary tasks~\cite{Kokkinos2016,XiTu15,Maninis2016,Man+16}.

\subsection{Training details}
\paragraph*{\textbf{Offline training}} 
The base CNN of our architecture~\cite{SiZi15} is pre-trained on ImageNet~\cite{Russakovsky2015} for image labeling,
which has proven to be a very good initialization to other tasks~\cite{LSD15,XiTu15,Kokkinos2016,Maninis2016,Har+15,Yan+16}.
Without further training, the network is not capable of performing segmentation, 
as illustrated in Figure~\ref{fig:overview} (1). We refer to this network as the \textit{``base network}.''

We therefore further train the network on the binary masks of the training set of DAVIS 2016,
to learn a generic notion of how to segment objects from their background, their usual shapes, etc.
We use Stochastic Gradient Descent (SGD) with momentum 0.9 for 50000
iterations.
We augment the data by mirroring and zooming in.
The learning rate is set to $10^{-8}$, and is gradually decreased.
After offline training, the network learns to segment foreground objects from the background,
as illustrated in Figure~\ref{fig:overview} (2).
We refer to this network as the \textit{``parent network}.''

\paragraph*{\textbf{Online training/testing}} With the parent network available, we can proceed to our main task (\textit{``test network}'' in Figure~\ref{fig:overview}):
Segmenting a particular entity in a video, given the image and the segmentation of the first frame.
We proceed by further training (fine-tuning) the parent network for the particular image/ground-truth pair, and then testing on the entire sequence, using the new weights.
The timing of our method is therefore affected by two times: the fine-tuning time (once per annotated mask)
and the segmentation of all frames (once per frame).
In the former we have a trade-off between quality and time: the more iterations we allow the technique to learn, the better results but the longer the user will have to wait for results.
The latter does not depend on the training time: 
\ours{} is able to segment each 480p frame ($480\times854$) in 130 ms.

\begin{figure}
\setlength{\fboxsep}{0pt}
\resizebox{\linewidth}{!}{%
      \fbox{\includegraphics[width=0.3\textwidth]{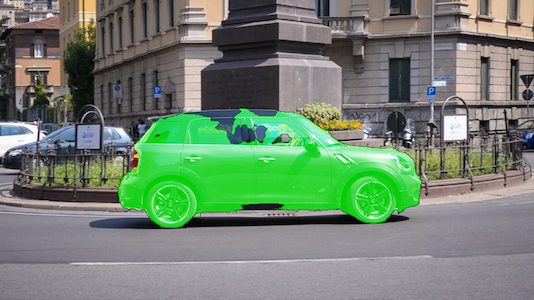}}
      \fbox{\includegraphics[width=0.3\textwidth]{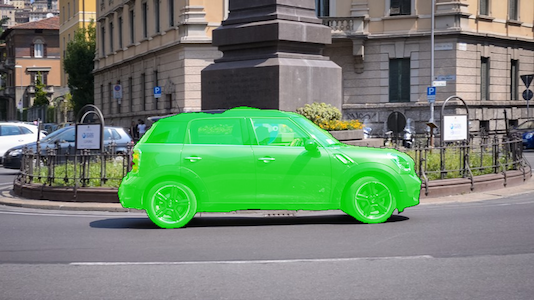}}
      }
\caption{\textbf{Qualitative evolution of the fine tuning}: Results at 10 seconds and 1 minute per sequence.}
\label{fig:learning}
\vspace{-4mm}
\end{figure}

Regarding the fine-tuning time, we present two different modes: One can either need to fine-tune online, by segmenting a frame and waiting for the results in the entire sequence, or offline, having access to the
object to segment beforehand.
Especially in the former mode, there is the need to control the amount of time dedicated to training: the more time allocated for fine-tuning, the more the user waits and the better the results are.
In order to explore this trade-off, in our experiments we train for a period between
10 seconds and 10 minutes per sequence.
Figure~\ref{fig:learning} shows a qualitative example of the evolution of the results quality depending on the time allowed for fine-tuning.
In the  experimental evaluation, Figure~\ref{fig:qual_vs_time} quantifies this evolution.

Ablation analysis shows that both offline and online training are crucial for good performance:
If we perform our online training directly from the base network (ImageNet model), the performance drops significantly. 
Only dropping the online training for a specific object (using the parent network directly) also yields a significantly worse performance, as already transpired from Figure~\ref{fig:overview}.

\begin{figure*}
\includegraphics[width=\linewidth]{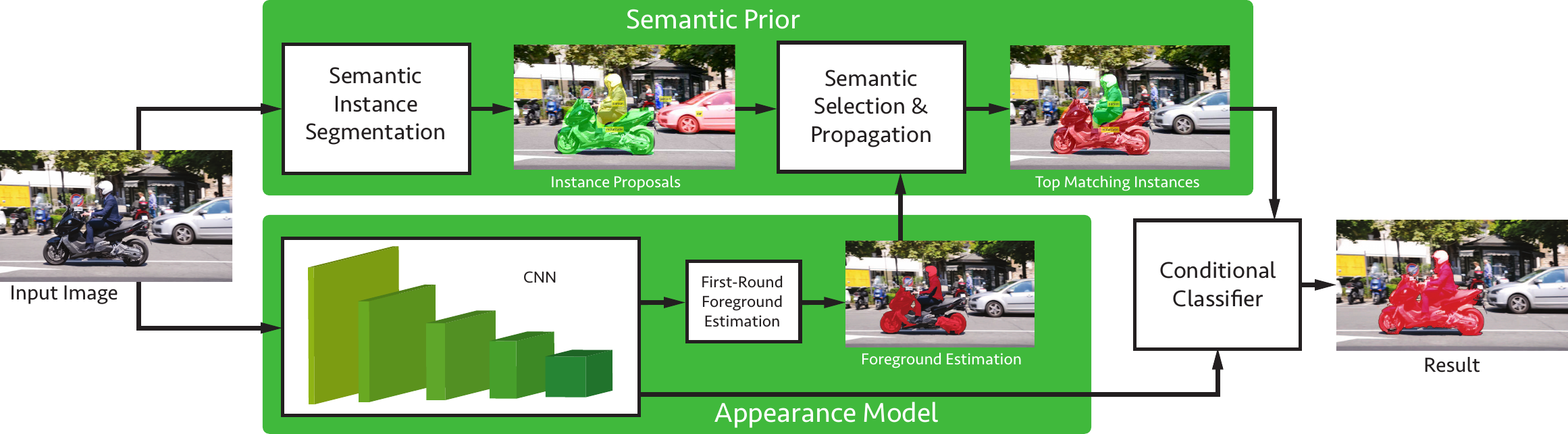}
\caption{\textbf{Network architecture overview}: Our network is composed of three major components: a base network as the feature extractor, and three classifiers built on top with shared features: a first-round foreground estimator to produce the semantic prior, and two conditional classifiers to model the appearance likelihood.}
\label{fig:overview_global}
\end{figure*}

\begin{figure*}
\includegraphics[width=\linewidth]{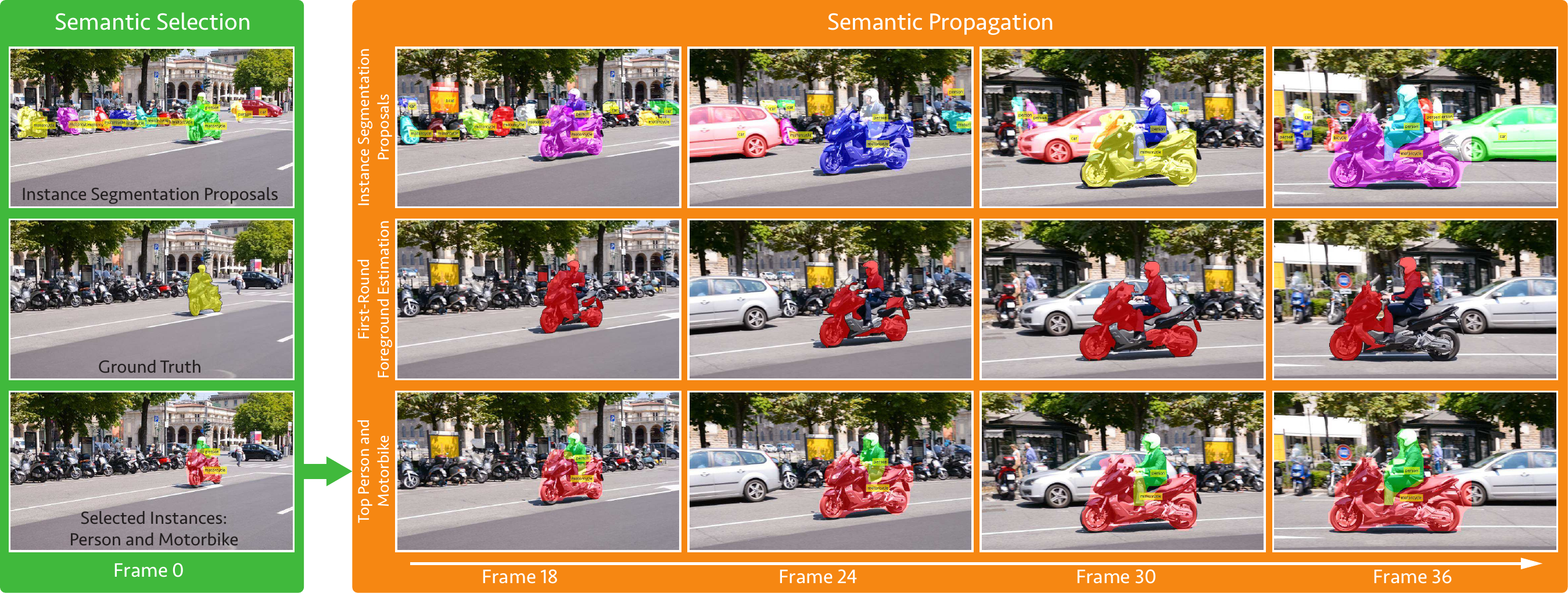}
\caption{\textbf{Semantic selection and propagation}: Illustrative example of the estimation of the semantics of the object from the first frame (semantic selection) and its propagation to the following frames (semantic propagation).}
\label{fig:overview_inst}
\end{figure*}

\subsection{Contour snapping}
In the field of image classification~\cite{Krizhevsky2012,SiZi15,He+16},
where our base network was designed and trained, spatial invariance is a design choice: no matter where an object appears in the image, the classification result should be the same. 
This is in contrast to the accurate localization of the object contours that we expect in (video) object segmentation.
Despite the use of skip connections~\cite{LSD15,Har+15,XiTu17,Man+16} to minimize the loss of spatial accuracy, we observe that \ours{}' segmentations have some room for improvement in terms of contour localization. 

To overcome this limitation, we propose a complementary CNN in a second branch that is trained to detect object contours. 
Figure~\ref{fig:two-stream} shows the global architecture: (1) shows the main foreground branch, where the foreground pixels are estimated; (2) shows the contour branch, which detects all contours in the scene (not only those of the foreground object).
This allows us to train offline, without the need to fine-tune on a specific example.
We used the exact same architecture in the two branches, but training for different losses.
We noticed that jointly training a network with shared layers for both tasks rather degrades the results thus we kept the computations for the two objectives uncorrelated.
This allows us to train the contour branch only offline and thus it does not affect the online timing.
Since there is need for high recall in the contours, we train on the PASCAL-Context~\cite{Mot+14} database, which provides contour annotations for the full scene of an image.

Once we have the estimated object contours, the boundary snapping step (Figure~\ref{fig:two-stream} (3)), consists of two different steps:
\begin{asparaenum}
\item[a)] \textbf{Superpixel snapping}: It computes superpixels that align to the computed contours (branch 2) by means of an Ultrametric Contour Map (UCM)~\cite{Arb+11,Pont-Tuset2016}, which we \textit{threshold} at a low strength. We then take a foreground mask (branch 1) and we select superpixels via majority voting (those that overlap with the foreground mask over 50\%) to form the final foreground segmentation.
\item[b)] \textbf{Contour recovery}: It recovers the very thin structures that are lost when snapping to superpixels. It enumerates the connected components of the foreground mask (branch 1), and then matches their contours to the detected contours in branch (2). The connected components whose contour matches the generic contours (branch 2) above a certain tolerance are added to the final result mask.
\end{asparaenum}

This refinement process results in a further boost in performance, and is fully modular, meaning that depending on the timing requirements one can choose not to use them, sacrificing accuracy for execution time; since the module comes with a small, yet avoidable computational overhead.
Please refer to the timing experiments (Figure~\ref{fig:qual_vs_time}) for a quantitative evaluation of this trade off: at which range of desired speeds one can afford to use contour snapping.

\section{Semantic Guidance (\oursnew)}
\label{sec:semantic}
The motivation behind semantic guidance is to improve the model we construct from the first frame with information about the category of the object and the number of instances, \eg we track two people and a motorbike.
We extract the semantic instance information from instance-aware semantic segmentation algorithms. We experiment with three top-performing methods: MNC~\cite{dai2016instance}, FCIS~\cite{Li+17}  and the most recent MaskRCNN~\cite{He+17}.
We modify the algorithm and the network architecture to select and propagate the specific instances we are interested in (Section~\ref{sec:sem_sel}), and then we adapt the network architecture to include these instance inside the CNN (Section~\ref{sec:cond_class}).
The global network overview is first presented in Section~\ref{sec:net_over}.

\subsection{Network Overview}
\label{sec:net_over}
Figure~\ref{fig:overview_global} illustrates the structure and workflow of the proposed semantic-aware network. Sharing the common base network (VGG) as the feature extractor, three pixel-wise classifiers are jointly learned. 

The first classifier, First-Round Foreground Estimation, is the original \ours{} head, which is purely appearance based, with no knowledge about the semantic segmentation source and produces the first foreground estimation. The result of that classifier and the information from an external semantic instance segmentation system are combined in the semantic selection and propagation steps (Section~\ref{sec:sem_sel}) to produce the top matching instances that we refer to as the semantic prior.

The two other classifiers inside the conditional classifier operate on both the features of the common base network and the semantic prior, and are dependent on each other: one is responsible for the pixels with a foreground prior, whereas the other for the background ones. Finally, the two sets of predictions are fused into the final prediction. See Section~\ref{sec:cond_class}.

\subsection{Semantic Selection and Semantic Propagation}
\label{sec:sem_sel}
We leverage a semantic instance segmentation algorithm as an input to estimate the semantics of the object to be segmented. Specifically, we choose MNC~\cite{dai2016instance}, FCIS~\cite{Li+17}, or MaskRCNN~\cite{He+17} as our input instance segmentation algorithms, and we use their publicly available implementations. We show that each of the improvements in instance segmentation is translated in a boost for the task of video object segmentation, which suggests that our method will be able to incorporate future improvements in the field.

The three instance semantic segmentation methods (MNC, FCIS, and MaskRCNN) are multi-stage networks that consist of three major components: shared convolutional layers, region proposal network (RPN), and region-of-interest(ROI)-wise classifiers. We use the available models which are pre-trained on PASCAL for the first one and on COCO for the other two. We note that our method is category agnostic, and the objects to segment do not necessarily need to be part of the PASCAL or COCO category vocabulary, as it will be shown in the experiments.

The output of the instance segmentation algorithm is given as a set of binary masks, together with their category, and their confidence of being a true object. We search for the object of interest inside the pool of most confident masks: our objective is to find a subset of masks with consistent semantics throughout the video as our semantic prior.

The process can be divided into two stages, namely \textit{semantic selection} and \textit{semantic propagation}.
Semantic selection happens in the first frame, where we select the masks that match the object according to the given ground-truth mask (please note that we are in a semi-supervised framework where the true mask of the first frame is given as input).
The number of instances and their categories are what we enforce to be consistent throughout the entire video. 
Figure~\ref{fig:overview_inst} depicts an example of both steps.
Semantic selection, on the left in {\color{green!60!black}green}, finds that we are interested in a motorbike plus a person (bottom), by overlapping the ground truth (middle) to the instance segmentation proposals (top). There are two cases where semantic selection may fail: a) the objects of interest are not part of the semantic vocabulary of the instance segmenter, and b) the wrong instances are selected by this step. Results show that our classifiers are robust to such failures, preserving high quality outputs in both cases. Thus, a fast greedy search for selecting the instances is sufficient to preserve high performance.

The semantic propagation stage (in {\color{orange}orange}) occurs at the following frames, where we propagate the semantic prior we estimated in the first frame to the following ones. No information from future frames is used in this stage.
The instance segmentation masks (first row), are filtered using the first-round foreground estimation from the \ours{} head (middle row), and the top matching person and motorbike from the pool are selected (bottom row). In cases that an instance of the selected classes does not overlap with the output of \ours{}, as in cases of occlusions and moving camera, we exclude the particular instance from the semantic prior, for the specific frame.

\subsection{Conditional Classifier}
\label{sec:cond_class}
Dense labeling using fully convolutional networks is commonly formulated as a per-pixel classification problem.
It can be therefore understood as a \textit{global} classifier sliding over the whole image,
and assigning either the foreground or background label to each pixel according to a \textit{monolithic} appearance model.
In this work, we want to incorporate the semantic prior to the final classification, which will be given as a mask of the most promising instance (or set of instances) in the current frame.

If semantic instance segmentation worked perfectly, we could directly select the best-matching instance to the appearance model, but in reality the results are far from perfect (as we will show in the experiments).
We can only, therefore, use the instance segmentation mask as a guidance, or a guess, of what the limits of that instance are, but we still need to perform a refinement step.
Our proposed solution to incorporate this mask but still keep the per-pixel classification is to train two classifiers and weigh them according to the confidence we have in that pixel being part of the instance or not.
We argue that using a single set of parameters for the whole image is suboptimal.

Formally, for each pixel $i$, we estimate its probability of being a foreground pixel given the image: $p(i|I)$.
The probability can be decomposed into the sum of $k$ conditional probabilities weighted by the prior:
\[p(i|I)=\sum_{k=1}^{K} p(i|I,k)\,p(k|I).\]
In our experiments, we use $K=2$, and we build two conditional classifiers, one focusing on the \textit{instance foreground} pixels, and the other focusing on the \textit{instance background} pixels.
The prior term $p(k|I)$ is estimated based on the instance segmentation output.
Specifically, a pixel relies more on the \textit{instance foreground} classifier if it is located within the instance segmentation mask; and more importance is given to the \textit{instance background} classifier if it falls out of the instance segmentation mask. In our experiments, we apply a Gaussian filter to spatially smooth the selected masks as our semantic prior.

\begin{figure}
\includegraphics[width=\linewidth]{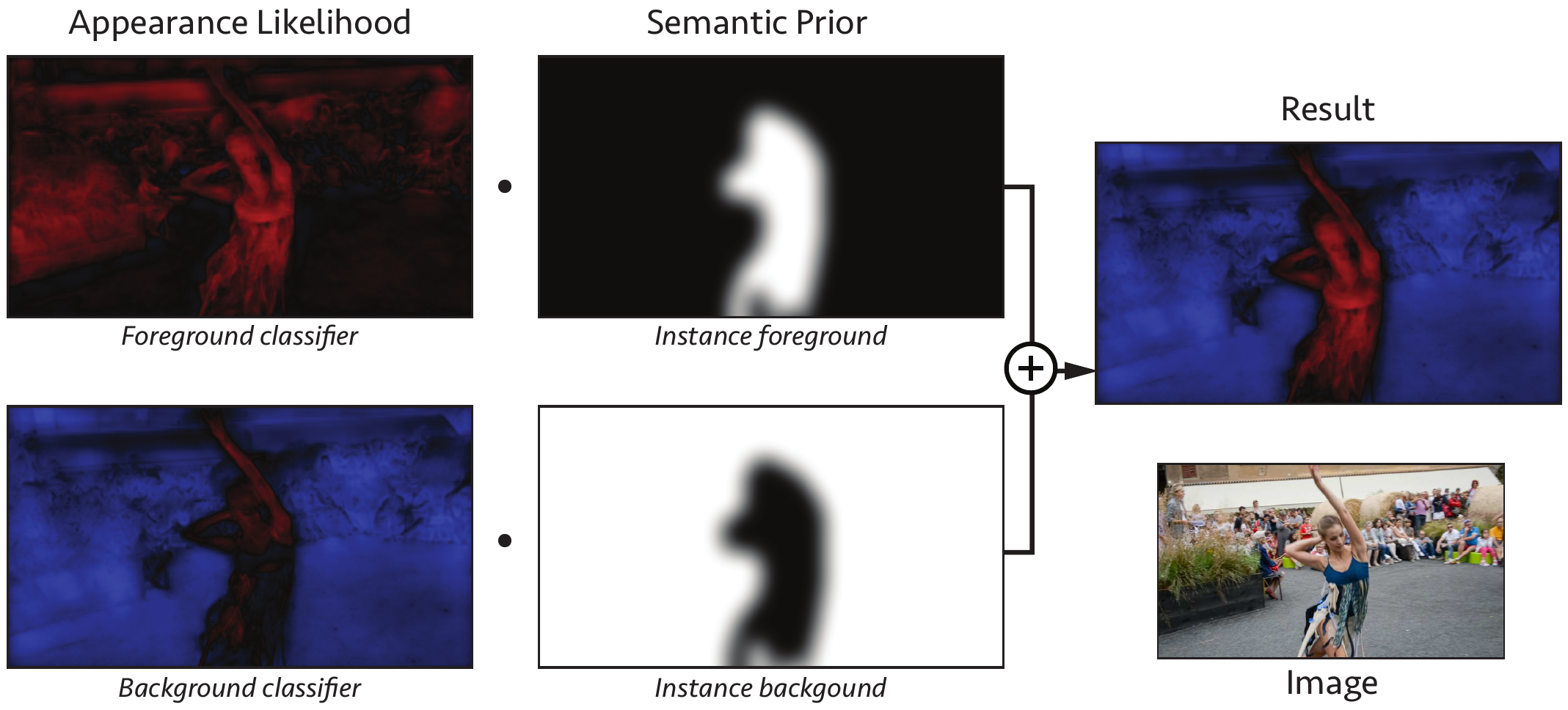}
\caption{\textbf{Forward pass of the conditional classifier layer}: {\color{red}Red} denotes foreground probability, and {\color{blue}blue} background probability. The output is the weighted sum of the two conditional classifier. }
\label{fig:classifier}
\end{figure}

\begin{figure*}
\centering
\begin{minipage}{0.19\linewidth}
\centering
\includegraphics[width=\linewidth]{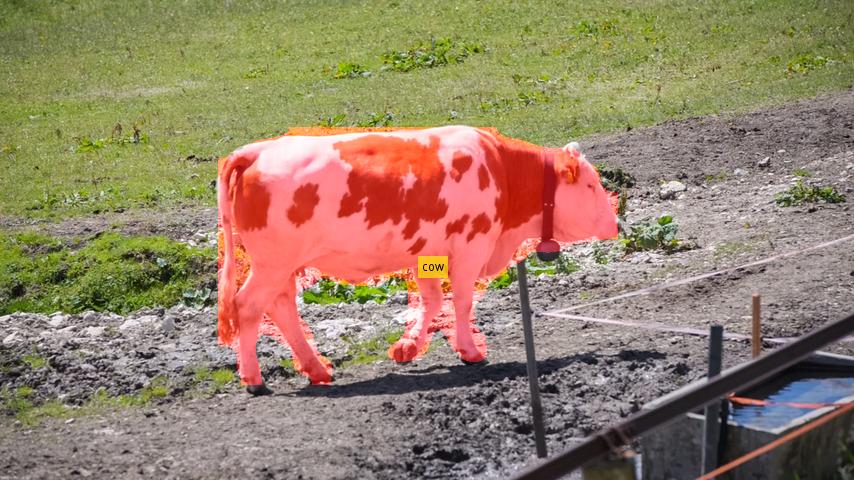}\\[-1pt]
\scriptsize(a) Cow
\end{minipage}
\hfill
\begin{minipage}{0.19\linewidth}
\centering
\includegraphics[width=\linewidth]{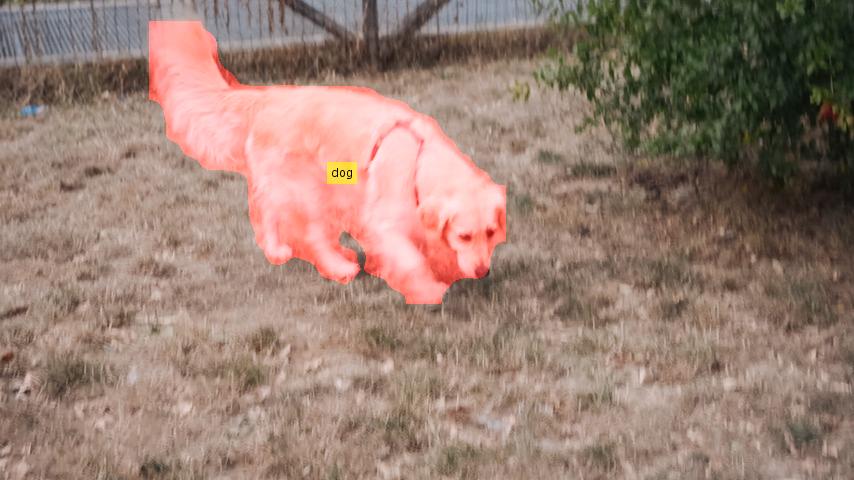}\\[-1pt]
\scriptsize(b) Dog
\end{minipage}
\hfill
\begin{minipage}{0.19\linewidth}
\centering
\includegraphics[width=\linewidth]{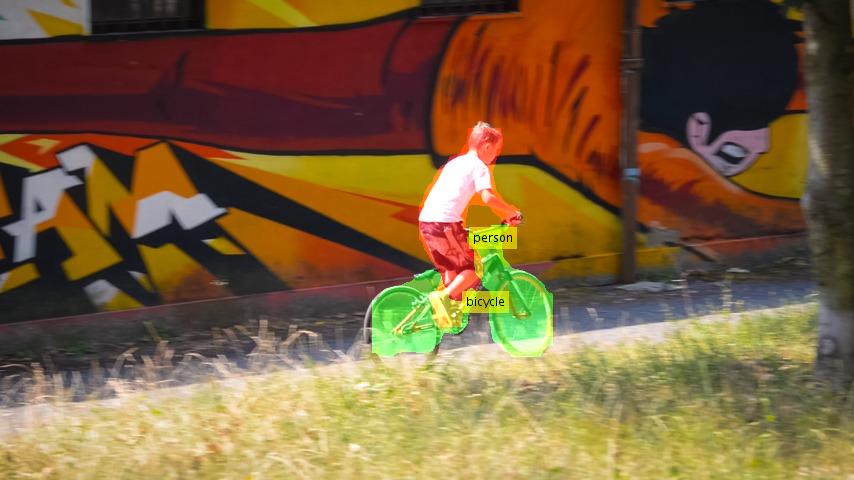}\\[-1pt]
\scriptsize(c) Person, bicycle
\end{minipage}
\hfill
\begin{minipage}{0.19\linewidth}
\centering
\includegraphics[width=\linewidth]{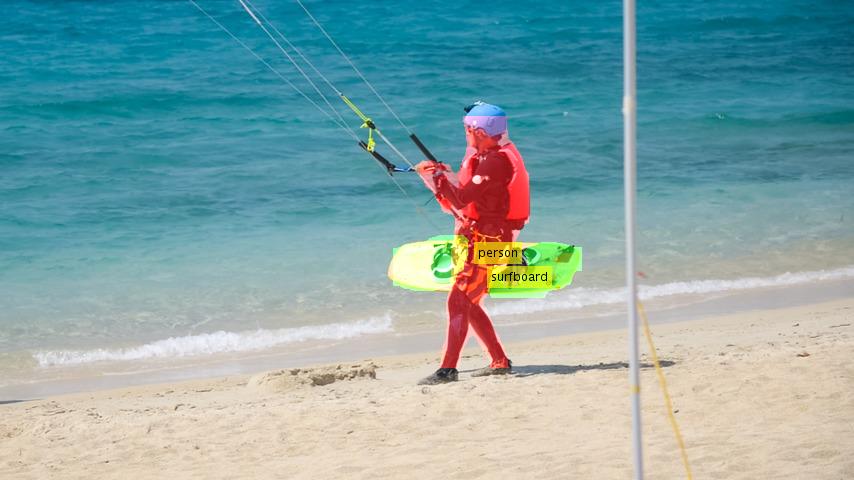}\\[-1pt]
\scriptsize(d) Person, surfboard
\end{minipage}
\hfill
\begin{minipage}{0.19\linewidth}
\centering
\includegraphics[width=\linewidth]{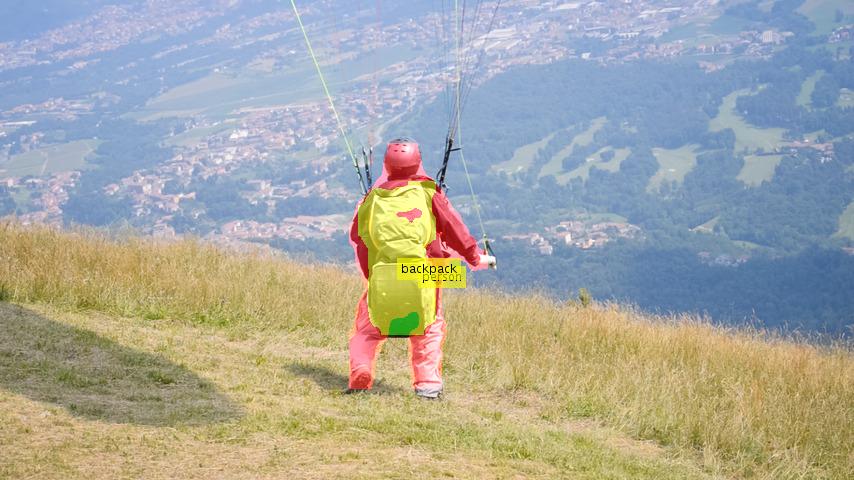}\\[-1pt]
\scriptsize(e) Person, backpack
\end{minipage}
\\[1.5mm]
\begin{minipage}{0.19\linewidth}
\centering
\includegraphics[width=\linewidth]{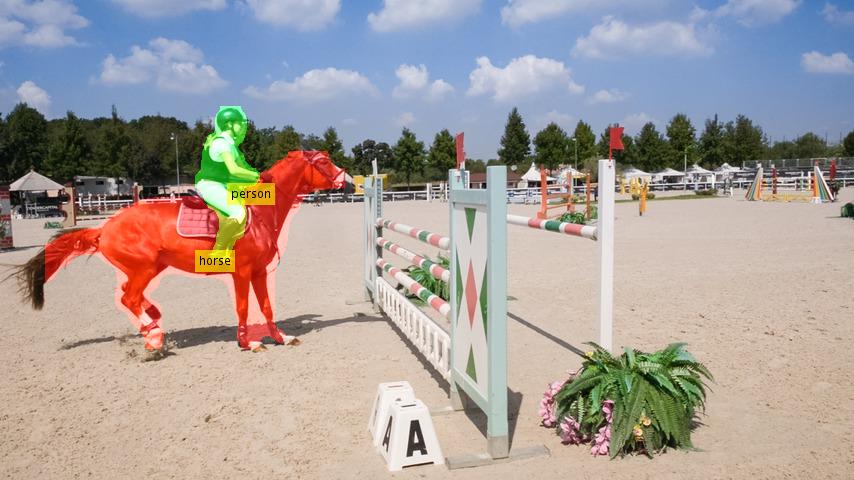}\\[-1pt]
\scriptsize(f) Person, horse
\end{minipage}
\hfill
\begin{minipage}{0.19\linewidth}
\centering
\includegraphics[width=\linewidth]{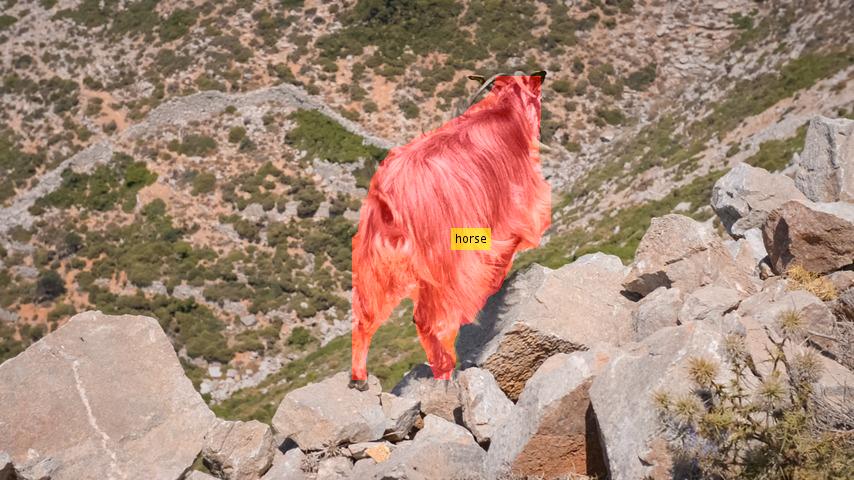}\\[-1pt]
\scriptsize(g) Horse
\end{minipage}
\hfill
\begin{minipage}{0.19\linewidth}
\centering
\includegraphics[width=\linewidth]{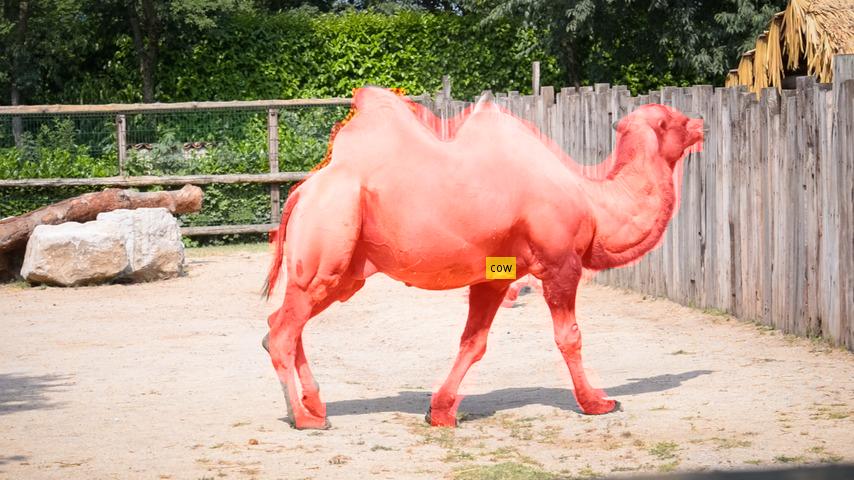}\\[-1pt]
\scriptsize(h) Cow
\end{minipage}
\hfill
\begin{minipage}{0.19\linewidth}
\centering
\includegraphics[width=\linewidth]{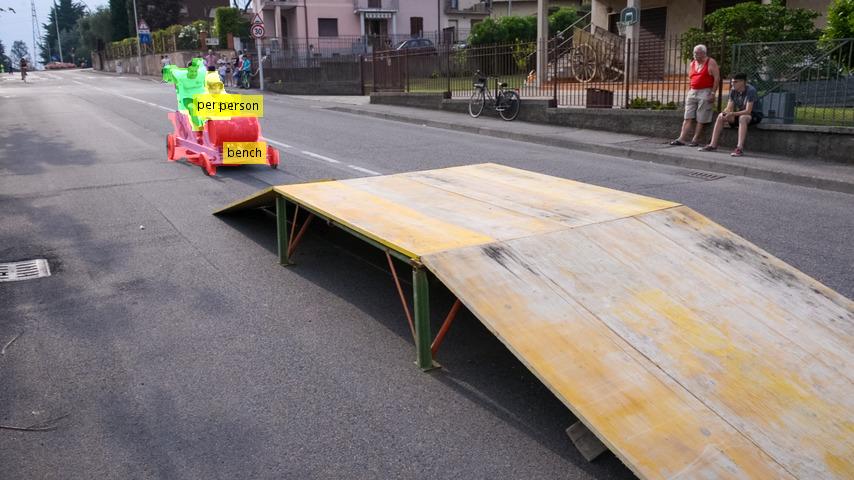}\\[-1pt]
\scriptsize(i) Person, person, bench
\end{minipage}
\hfill
\begin{minipage}{0.19\linewidth}
\centering
\includegraphics[width=\linewidth]{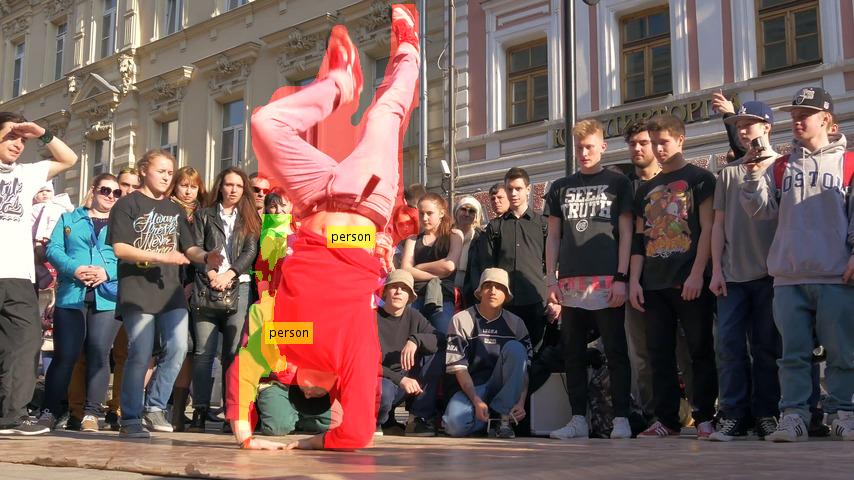}\\[-1pt]
\scriptsize(j) Person, person
\end{minipage}
\\[0.5mm]
\caption{\textbf{Semantic selection evaluation}: Semantic instances selected by the semantic selection step, with its category overlaid. We observe that in some cases either the semantic labels (h-i) or the number of instances (j) is incorrect. The final results, however, are robust to such mistakes.}
\label{fig:sem_sel}
\end{figure*}

The conditional classifier is implemented as a layer which can be easily integrated in the network in a end-to-end trainable manner. The layer takes two prediction maps $f_1$ and $f_2$ and the weight maps $p(k|I)$ which come from the semantic selection. Without loss of generality, we will assume that $k=1$ corresponds to the foreground of the semantic prior. For convenience, we set $w=p(k\!=\!1|I)$, and in our case $1-w=p(k\!=\!2|I)$ (background prior).
The inference process is illustrated in Figure~\ref{fig:classifier}, where each input element is multiplied by its corresponding weight from the weight map,
then summed with the corresponding element in the other map:
\begin{equation}
\label{eq:cond}
f_{out}(x,y) = w(x,y)\,f_{1}(x,y) + \big(1\!-\!w(x,y)\big)\,f_{2}(x,y).
\end{equation}

In Equation~\ref{eq:cond}, $x$ and $y$ represent the horizontal and vertical pixel location on a frame. This equation suggest that the decision for the pixels near the selected instances are made by the instance foreground classifier ($f_{1}(x,y)$), whereas the instance background classifier ($f_{2}(x,y)$) decides for the rest of the pixels.

Similarly, in the back-propagation step, the gradient from the top $g_{top}$ is propagated to the
two parts according to the weight map:
\begin{align*}
g_1(x,y)&= w(x,y)\,g_{top}(x,y)\\
g_2(x,y)&=\big(1-w(x,y)\big)\,g_{top}(x,y).
\end{align*}

The conditional classifier is necessary to incorporate the semantic prior information, in order to make softer decisions. Techniques that can be used as alternatives incorporating only a single classifier, such as masking of the features by the semantic prior, lead to hard decisions guided by the semantics, unable to recover in regions where they are wrong. For example, in Figure~\ref{fig:classifier}, the left hand of the dancer is not detected by the semantic prior, and it will be immediately classified as background in the case of feature masking. The background classifier of our proposed method, however, is able to recover the region, correctly classifying it as a foreground.

\subsection{Training and Inference}
We follow the same ideas as \ours{} to train and test the network, every step enriched with the semantic selection and propagation steps. The parent network is trained using semantic instances that overlap with the ground-truth masks of the DAVIS 2016 training set. Similarly, during online fine-tuning we use the label of the first frame, as well as the outputs of the OSVOS head for the next ones. As was done before, each frame is processed independently of the others. As shown in the experiments, the plug-in of the instance segmentation module dramatically improves the quality of the final segmentation.

\section{Experimental Validation}

\paragraph*{\textbf{Experimental Setup}}
We will mainly work on the DAVIS 2016 database~\cite{Perazzi2016}, using their proposed metrics:
region similarity (intersection over union $\J$),
contour accuracy ($\F$ measure), and temporal instability ($\T$).
The dataset contains 50 full-HD annotated video sequences, 30 in the training set and 20 in the validation set.
All our results will be trained on the former, evaluated on the latter.
As a global comparison metric we will use the mean between $\J$ and $\F$, as proposed in the DAVIS 2017 challenge~\cite{Pont-Tuset2017}.

\begin{table*}
\setlength{\tabcolsep}{4pt} 
\centering
\footnotesize
\rowcolors{1}{white}{rowblue}
\resizebox{0.65\linewidth}{!}{%
\sisetup{detect-all=true}
\begin{tabular}{llS[table-format=2.1]S[table-format=2.1]@{\hspace{1.5mm}}>{\fontsize{6}{6}}S[table-format=2.1]S[table-format=2.1]@{\hspace{1.5mm}}>{\fontsize{6}{6}}S[table-format=2.1]S[table-format=2.1]@{\hspace{1.5mm}}>{\fontsize{6}{6}}S[table-format=2.1]S[table-format=2.1]@{\hspace{1.5mm}}>{\fontsize{6}{6}}S[table-format=2.1]S[table-format=2.1]@{\hspace{1.5mm}}>{\fontsize{6}{6}}S[table-format=2.1]}
\toprule
\multicolumn{2}{c}{Measure} & \si{ImageNet} & \multicolumn{2}{c}{\si{+OneShot}} & \multicolumn{2}{c}{\si{+Parent}} & \multicolumn{2}{c}{\si{+Semantics}} & \multicolumn{2}{c}{\si{+Superpixels}} & \multicolumn{2}{c}{\si{+Contours}} \\
\midrule\cellcolor{rowblue}$\mathcal{J}\&\mathcal{F}$  & Mean $\mathcal{M} \uparrow$   & 18.9 &          65.6 &\itshape \color{blue}46.7 &          77.8 &\itshape \color{blue}12.1 &          86.1 &\itshape \color{blue}8.4 &          85.4 &\itshape \color{red}0.7 &\bfseries 86.5 &\itshape \color{blue}1.1 \\
\hline
                                 & Mean $\mathcal{M} \uparrow$     & 17.6 &          64.6 &\itshape \color{blue}47.0 &          77.4 &\itshape \color{blue}12.8 &          85.0 &\itshape \color{blue}7.6 &          85.5 &\itshape \color{blue}0.5 &\bfseries 85.6 &\itshape \color{blue}0.1 \\
\cellcolor{white}$\mathcal{J}$ & Recall $\mathcal{O} \uparrow$   & 2.3 &          70.5 &\itshape \color{blue}68.2 &          91.0 &\itshape \color{blue}20.5 &          96.7 &\itshape \color{blue}5.7 &          96.5 &\itshape \color{red}0.2 &\bfseries 96.8 &\itshape \color{blue}0.3 \\
                                 & Decay $\mathcal{D} \downarrow$  &\bfseries 1.8 &          27.8 &\itshape \color{red}26.0 &          17.4 &\itshape \color{blue}10.4 &          7.2 &\itshape \color{blue}10.2 &          5.9 &\itshape \color{blue}1.4 &          5.5 &\itshape \color{blue}0.3 \\
\hline
                                 & Mean $\mathcal{M} \uparrow$     & 20.3 &          66.7 &\itshape \color{blue}46.4 &          78.1 &\itshape \color{blue}11.4 &          87.3 &\itshape \color{blue}9.2 &          85.3 &\itshape \color{red}2.0 &\bfseries 87.5 &\itshape \color{blue}2.2 \\
\cellcolor{rowblue}$\mathcal{F}$ & Recall $\mathcal{O} \uparrow$   & 2.4 &          74.4 &\itshape \color{blue}72.0 &          92.0 &\itshape \color{blue}17.6 &\bfseries 95.9 &\itshape \color{blue}3.9 &          94.1 &\itshape \color{red}1.8 &          95.9 &\itshape \color{blue}1.8 \\
                                 & Decay $\mathcal{D} \downarrow$  &\bfseries 2.4 &          26.4 &\itshape \color{red}24.0 &          19.4 &\itshape \color{blue}7.0 &          9.3 &\itshape \color{blue}10.1 &          6.8 &\itshape \color{blue}2.5 &          8.2 &\itshape \color{red}1.5 \\
\hline
\cellcolor{white}$\mathcal{T}$ & Mean $\mathcal{M} \downarrow$   & 46.0 &          60.9 &\itshape \color{red}14.9 &          33.5 &\itshape \color{blue}27.4 &\bfseries 20.2 &\itshape \color{blue}13.3 &          25.1 &\itshape \color{red}4.9 &          21.7 &\itshape \color{blue}3.4 \\
\bottomrule
\end{tabular}
}\\[1.5mm]
\caption{\textbf{Ablation study on DAVIS 2016}: From a network pretrained on ImageNet, all improvement steps to the proposed \oursnew{} (right-most column).
Numbers in italics show how much the results improve (in blue) or worsen (in red) in that metric with respect to the previous column.}
\label{tab:ablation}
\end{table*}%

We compare against a large body of very recent semi-supervised state-of-the-art techniques (\nonavos~\cite{\onavos}, \nmsk~\cite{\msk}, \nctn~\cite{\ctn}, \nvpn~\cite{\vpn}, \nofl~\cite{\ofl}, \nbvs~\cite{\bvs}, and \nfcp~\cite{\fcp}) using the pre-computed results provided by the respective authors.
For context, we also add the results of the latest unsupervised techniques (\narp~\cite{\arp}, \nfseg~\cite{\fseg}, \nlmp~\cite{\lmp}, \nfst~\cite{\fst}, \nnlc~\cite{\nlc}, \nmsg~\cite{\msg}).

Moreover, we perform experiments on DAVIS 2017 which contains videos with multiple objects. We compute the results on the \textit{test-dev} set using the submission website provided by the organizers of the challenge. We compare against \nonavos~\cite{\onavos}, its submission to the DAVIS 2017 challenge which achieves the fifth place~\cite{DAVIS2017-5th} and the other top-performing methods of the challenge~\cite{DAVIS2017-1st, DAVIS2017-2nd, DAVIS2017-3rd, DAVIS2017-4th}.

For completeness, we also experiment on the Youtube-objects dataset~\cite{Prest2012,Jain2014}
against those techniques with public segmentation results (\nonavos~\cite{\onavos}, \nosvos~\cite{\osvos}, \nmsk~\cite{\msk}, \nofl~\cite{\ofl}, \nbvs~\cite{\bvs}).
We do not take pre-computed evaluation results directly from the paper tables because the benchmarking algorithm is not consistent among the different authors.

\begin{table}
\setlength{\tabcolsep}{2pt} 
\center
\footnotesize
\rowcolors{3}{rowblue}{white}
\resizebox{\linewidth}{!}{%
\sisetup{detect-all=true}
\begin{tabular}{ll@{\hspace{3mm}}S[table-format=2.1]S[table-format=2.1]@{\hspace{3mm}}S[table-format=2.1]S[table-format=2.1]@{\hspace{3mm}}S[table-format=2.1]S[table-format=2.1]@{\hspace{3mm}}S[table-format=2.1]}
\toprule
\multicolumn{2}{c}{} & \multicolumn{2}{c}{MNC\rule{2mm}{0mm}} & \multicolumn{2}{c}{FCIS\rule{2mm}{0mm}} & \multicolumn{2}{c}{Mask-RCNN\rule{2mm}{0mm}} & \\
\multicolumn{2}{c}{Measure} & \si{Automatic} & \si{Oracle} & \si{Automatic} & \si{Oracle}  & \si{Automatic} & \si{Oracle} & \si{\oursnew} \\
\cmidrule(lr){1-2} \cmidrule(lr){3-9}
\cellcolor{rowblue}$\mathcal{J}\&\mathcal{F}$  & $\mathcal{M} \uparrow$   &          63.7 &          81.5 &          73.1 &          75.1 &          82.4 &          82.6 &\bfseries 86.5 \\
\hline
                                 & $\mathcal{M} \uparrow$     &          68.9 &          81.3 &          74.3 &          76.4 &          82.6 &          82.8 &\bfseries 85.6 \\
\cellcolor{white}$\mathcal{J}$ & $\mathcal{O} \uparrow$   &          85.5 &          95.8 &          88.4 &          92.0 &          93.7 &          94.3 &\bfseries 96.8 \\
                                 & $\mathcal{D} \downarrow$  &          3.3 &          8.4 &          1.9 &\bfseries 1.8 &          3.0 &          2.2 &          5.5 \\
\hline
                                 & $\mathcal{M} \uparrow$     &          58.5 &          81.6 &          71.9 &          73.7 &          82.2 &          82.3 &\bfseries 87.5 \\
\cellcolor{rowblue}$\mathcal{F}$ & $\mathcal{O} \uparrow$   &          63.0 &          93.3 &          82.8 &          87.2 &          88.8 &          89.5 &\bfseries 95.9 \\
                                 & $\mathcal{D} \downarrow$  &          3.0 &          13.6 &          3.0 &          3.2 &          3.4 &\bfseries 2.7 &          8.2 \\
\hline
\cellcolor{white}$\mathcal{T}$ & $\mathcal{M} \downarrow$   &          30.5 &          28.4 &          24.8 &          23.9 &\bfseries 17.4 &          17.5 &          21.7 \\
\bottomrule
\end{tabular}
}
\vspace{2mm}
\caption{\label{tab:sem_prop}
\textbf{Semantic propagation}: Comparing the automatic selection of instances against an oracle and our 
final result.}
\vspace{0mm}
\end{table}

\begin{table*}
\setlength{\tabcolsep}{4pt} 
\centering
\footnotesize
\rowcolors{5}{white}{rowblue}
\resizebox{\textwidth}{!}{%
\sisetup{detect-weight=true}
%
\begin{tabular}{llS[table-format=2.1]S[table-format=2.1]S[table-format=2.1]S[table-format=2.1]S[table-format=2.1]S[table-format=2.1]S[table-format=2.1]S[table-format=2.1]S[table-format=2.1]S[table-format=2.1]S[table-format=2.1]S[table-format=2.1]S[table-format=2.1]S[table-format=2.1]S[table-format=2.1]S[table-format=2.1]}
\toprule
	 & & \multicolumn{9}{c}{Semi-Supervised} & \multicolumn{6}{c}{Unsupervised} & \multicolumn{1}{c}{Bound} \\
	 \cmidrule(lr){3-11} \cmidrule(lr){12-17} \cmidrule(lr){18-18}
\multicolumn{2}{c}{Measure} & \si{OSVOS^S} & \si{\nonavos} & \si{\nosvos} & \si{\nmsk} & \si{\nctn} & \si{\nvpn} & \si{\nofl} & \si{\nbvs} & \si{\nfcp} & \si{\narp} & \si{\nfseg} & \si{\nlmp} & \si{\nnlc} & \si{\nfst} & \si{\nmsg} & \si{\ncobspub} \\
\cmidrule(lr){1-2} \cmidrule(lr){3-11} \cmidrule(lr){12-17} \cmidrule(lr){18-18}
\cellcolor{rowblue}$\mathcal{J}\&\mathcal{F}$  & Mean $\mathcal{M} \uparrow$   &\bfseries 86.5 &          85.5 &          80.2 &          77.5 &          71.4 &          67.8 &          65.7 &          59.4 &          53.8 &\bfseries 73.4 &          68.0 &          67.9 &          53.7 &          53.4 &          52.1 &\bfseries 86.8 \\
\hline
                                 & Mean $\mathcal{M} \uparrow$     &          85.6 &\bfseries 86.1 &          79.8 &          79.7 &          73.5 &          70.2 &          68.0 &          60.0 &          58.4 &\bfseries 76.2 &          70.7 &          70.0 &          55.1 &          55.8 &          53.3 &\bfseries 86.5 \\
\cellcolor{white}$\mathcal{J}$ & Recall $\mathcal{O} \uparrow$   &\bfseries 96.8 &          96.1 &          93.6 &          93.1 &          87.4 &          82.3 &          75.6 &          66.9 &          71.5 &\bfseries 91.1 &          83.5 &          85.0 &          55.8 &          64.9 &          61.6 &\bfseries 96.5 \\
                                 & Decay $\mathcal{D} \downarrow$  &          5.5 &          5.2 &          14.9 &          8.9 &          15.6 &          12.4 &          26.4 &          28.9 &\bfseries -2.0 &          7.0 &          1.5 &          1.3 &          12.6 &\bfseries -0.0 &          2.4 &\bfseries 2.8 \\
\hline
                                 & Mean $\mathcal{M} \uparrow$     &\bfseries 87.5 &          84.9 &          80.6 &          75.4 &          69.3 &          65.5 &          63.4 &          58.8 &          49.2 &\bfseries 70.6 &          65.3 &          65.9 &          52.3 &          51.1 &          50.8 &\bfseries 87.1 \\
\cellcolor{rowblue}$\mathcal{F}$ & Recall $\mathcal{O} \uparrow$   &\bfseries 95.9 &          89.7 &          92.6 &          87.1 &          79.6 &          69.0 &          70.4 &          67.9 &          49.5 &\bfseries 83.5 &          73.8 &          79.2 &          51.9 &          51.6 &          60.0 &\bfseries 92.4 \\
                                 & Decay $\mathcal{D} \downarrow$  &          8.2 &          5.8 &          15.0 &          9.0 &          12.9 &          14.4 &          27.2 &          21.3 &\bfseries -1.1 &          7.9 &\bfseries 1.8 &          2.5 &          11.4 &          2.9 &          5.1 &\bfseries 2.3 \\
\hline
\cellcolor{white}$\mathcal{T}$ & Mean $\mathcal{M} \downarrow$   &          21.7 &\bfseries 19.0 &          37.8 &          21.8 &          22.0 &          32.4 &          22.2 &          34.7 &          30.6 &          39.3 &          32.8 &          57.2 &          42.5 &          36.6 &\bfseries 30.1 &\bfseries 27.9 \\

\hline
\multicolumn{2}{c}{Training Images} & ~2.3\,k + {\it83\,k}$^\S$ & ~87\,k$^\S$ & ~2.3\,k$^\S$ & ~11\,k$^\S$ & ~11.4\,k$^\S$ & ~2.3\,k$^\S$ & 1 & 1 & 1 & \textemdash & \textemdash & \textemdash & \textemdash & \textemdash & \textemdash & \textemdash \\
\bottomrule
\end{tabular}
}
\vspace{1mm}
\caption{\label{tab:evaltable}\textbf{DAVIS 2016 Validation}: \oursnew{} versus the state of the art (both semi- and un-supervised, and a practical bound). For the number of images, we count those datasets that have some form of segmentation (instance or semantic), and we mark the models pre-trained on Imagenet with $^\S$. $k$ stands for thousands. The number of images in italics is not directly used to train for the task of video object segmentation, but to train the auxiliary semantic instance segmentation network used by \oursnew.}
\vspace{-5mm}
\end{table*}%

\begin{figure*}
\pgfplotstableread{data/per_seq_mean_JF.txt}\perseqdata
\mbox{%
\begin{minipage}{0.88\textwidth}
  \resizebox{\textwidth}{!}{%
    \begin{tikzpicture}
        \begin{axis}[set layers,width=1.4\textwidth,height=0.35\textwidth,
                ybar=0pt,bar width=0.10,
                grid=both,
				grid style=dotted,
                minor ytick={0,0.05,...,1.1},
    			ytick={0,0.1,...,1.1},
			    yticklabels={0,.1,.2,.3,.4,.5,.6,.7,.8,.9,1},
                ymin=0, ymax=1,
                xtick = data, x tick label style={rotate=20,anchor=north east,xshift=7pt,yshift=5pt},
                xticklabels from table={\perseqdata}{Seq},
                major x tick style = transparent,
                enlarge x limits=0.03,
                font=\scriptsize,
                ]
             \addplot[draw opacity=0,fill=Set2-8-1,mark=none,legend image post style={yshift=-0.1em}] table[x expr=\coordindex,y=ONAVOS]{\perseqdata};
            \label{fig:perseq:onavos}
            \addplot[draw opacity=0,fill=Set2-8-2,mark=none,legend image post style={yshift=-0.1em}] table[x expr=\coordindex,y=OSVOS]{\perseqdata};
            \label{fig:perseq:osvos}
            \addplot[draw opacity=0,fill=Set2-8-3,mark=none,legend image post style={yshift=-0.1em}] table[x expr=\coordindex,y=MSK]{\perseqdata};
            \label{fig:perseq:msk}
            \addplot[draw opacity=0,fill=Set2-8-4,mark=none,legend image post style={yshift=-0.1em}] table[x expr=\coordindex,y=CTN]{\perseqdata};
            \label{fig:perseq:ctn}
            \addplot[draw opacity=0,fill=Set2-8-5,mark=none,legend image post style={yshift=-0.1em}] table[x expr=\coordindex,y=VPN]{\perseqdata};
            \label{fig:perseq:vpn}
            \addplot[draw opacity=0,fill=Set2-8-6,mark=none,legend image post style={yshift=-0.1em}] table[x expr=\coordindex,y=OFL]{\perseqdata};
            \label{fig:perseq:ofl}
            \addplot[draw opacity=0,fill=Set2-8-7,mark=none,legend image post style={yshift=-0.1em}] table[x expr=\coordindex,y=BVS]{\perseqdata};
			\label{fig:perseq:bvs}
			\addplot[draw opacity=0,fill=Set2-8-8,mark=none,legend image post style={yshift=-0.1em}] table[x expr=\coordindex,y=FCP]{\perseqdata};
			\label{fig:perseq:fcp}
            \addplot[blue,sharp plot,update limits=false,mark=*,mark size=1,line width=1.2pt, legend image post style={yshift=-0.4em}] table[x expr=\coordindex,y=OSVOSS]{\perseqdata};
            \label{fig:perseq:osvoss}
            
            \addplot[black,sharp plot,update limits=false] coordinates{(-0.5,0) (20.5,0)};
        \end{axis}
    \end{tikzpicture}
    }
    \end{minipage}
    \hspace{0mm}
    \begin{minipage}{0.07\textwidth}
    \scriptsize
    \begin{tabular}{@{}l@{\hspace{1.5mm}}l}
    \ref{fig:perseq:osvoss}& \oursnew{} (Ours)\\
    \ref{fig:perseq:onavos} &\nonavos~\cite{\onavos}\\
    \ref{fig:perseq:osvos} &\nosvos{} (Ours)\\
    \ref{fig:perseq:msk} &\nmsk~\cite{\msk}\\
    \ref{fig:perseq:ctn} &\nctn~\cite{\ctn}\\
    \ref{fig:perseq:vpn} &\nvpn~\cite{\vpn}\\
    \ref{fig:perseq:ofl} &\nofl~\cite{\ofl}\\
    \ref{fig:perseq:bvs} &\nbvs~\cite{\bvs}\\
    \ref{fig:perseq:fcp} &\nfcp~\cite{\fcp}
   \end{tabular}
    \end{minipage}
    }
    \vspace{-3mm}
    \caption{\label{fig:perseq}\textbf{DAVIS 2016 Validation}: Per-sequence results of mean region similarity and contour accuracy ($\J\&\F$).}
\end{figure*}
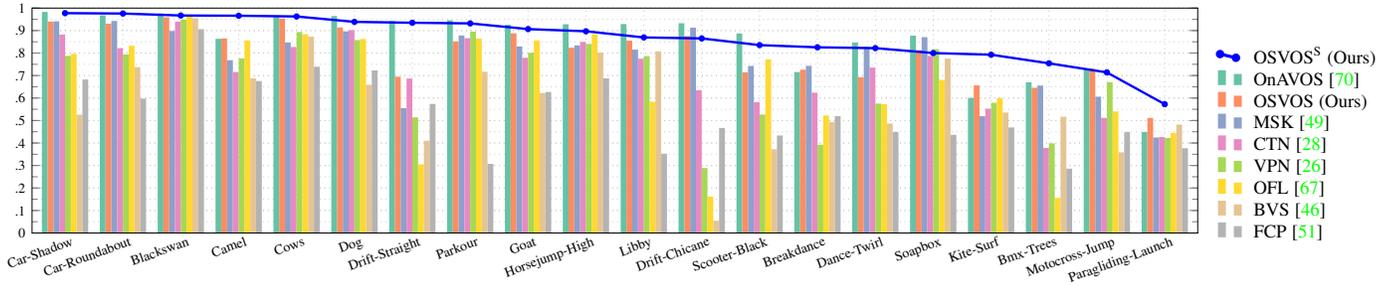

\paragraph*{\textbf{Ablation Study}}
Table~\ref{tab:ablation} shows how much each of the improvements presented builds up to the final result. We start by evaluating the network using only ImageNet pre-trained weights, before including any further training to the pipeline. The results in terms of segmentation ($\J\&\F=18.9\%$) are completely random (as visually shown in Figure~\ref{fig:overview}).
Fine-tuning on the mask of the first frame already boosts the results to competitive levels (+OneShot).
By pre-training the parent model, we allow fine-tuning to start from a much more meaningful set of weights, from a problem closer to the final one, so performance increases by 12\% (+Parent).
Adding semantics and the conditional classifier (+Semantics) plays an important role
both in terms of regions and contours ($\J\&\F$), but especially on temporal stability ($\T$).
Snapping to superpixels (+Superpixels) and recovering the contours (+Contours) improve the results around half a point overall, the former especially in terms of $\J$,
the latter in terms of $\F$, as it stands to reason.

\paragraph*{\textbf{Semantic Selection and Propagation}}
Figure~\ref{fig:sem_sel} qualitatively evaluates the semantic-selection algorithm:
it displays the selected semantic instances on the first frame of eight videos.
Examples (a) and (b) show correct detections in terms of category when a single instance is present.
Results (c) to (f) show that the algorithm works also in terms of the quantity of instances when more than one of them is needed.
Images (g) to (i) display cases where the category of the object is not present in MS COCO~\cite{Lin2014} (on which the instance segmentation algorithm was trained),
so the \textit{closest semantic} match is used instead.
Please note that the precise category is not needed for our algorithm to work, as long as that category is consistent throughout the video (\eg as long as the camel is always detected as a cow).
Last image (j) shows a failure case where two persons are detected when just a single one
(albeit upside down) is present, but the algorithm is afterwards robust to this mistake.

Once the semantic selection is done on the first frame, the information is propagated throughout the video.
Table~\ref{tab:sem_prop} quantitatively evaluates this step by comparing our automatic selection of instances
against an oracle that selects the best instance in each frame independently.
We use three different instance segmentation algorithms (MNC~\cite{dai2016instance}, FCIS~\cite{Dai2016a} and MaskRCNN~\cite{He+17}).
The results show that in all cases our automatic selection gets very close to the oracle selection (best possible instance), so 
we are not losing much quality in this step; and this is so in all instance segmentation algorithms,
showing that we are robust to the particular algorithm used and so we will be able to incorporate 
future improvements in this front.
The last column shows our final result, which significantly improves the oracle selection,
so instance segmentation alone is not enough, as already pointed out in previous sections. For the rest of the paper, we refer to \oursnew{} as the \oursnew{}-MaskRCNN variant of our method.

\paragraph*{\textbf{Comparison to State of the Art in DAVIS 2016}}
Table~\ref{tab:evaltable} shows the comparison of \oursnew{} and \ours{} against a large set of
very recent video segmentation algorithms, semi-supervised (using the first segmented frame as input) and unsupervised (only the raw video as input).
Apart from the standard metrics of DAVIS 2016~\cite{Perazzi2016}, we also add the most recent mean between $\J$ and $\F$,
as used in the 2017 DAVIS Challenge~\cite{Pont-Tuset2017}.

\oursnew{} is the best performing technique overall, one point above the second semi-supervised technique and 12.6 points above the best unsupervised one.
Last column shows the best result one could obtain from picking superpixels from COB~\cite{Man+17,Maninis2016},
a state-of-the-art generic image segmentation algorithm, at a very fine scale. We select the superpixels by snapping the ground-truth masks to them, thus creating a very strong bound.
\oursnew{} is only 0.3 points below the value of this oracle, further highlighting the outstanding quality of our results.

Next, we break down the performance on DAVIS 2016 per sequence. Figure~\ref{fig:perseq} shows the previous
state-of-the-art techniques in bars, and \oursnew{} using a line; sorted by the \textit{difficulty}
of the sequence for our technique.
We see that we outperform the majority of algorithms in the majority of sequences, especially so in the 
more challenging ones (\eg Kite-Surf, Bmx-Trees).
Please also note that \oursnew{} results are above 70\% in all but one sequence and above 80\% in all but three, which highlights the robustness of the approach.

Table~\ref{tab:evaltableattr} shows the per-attribute comparison in DAVIS 2016, that is,
the mean results on a subset of sequences where a certain challenging attribute is present (\eg camera shake or occlusions).
The increase/decrease of performance when each attribute is not present (small positive/negative numbers in italics)
is significantly low, which shows that \oursnew{} is also very robust to the different challenges.

\begin{table}
\setlength{\tabcolsep}{4pt} 
\center
\footnotesize
\rowcolors{1}{white}{rowblue}
\resizebox{\linewidth}{!}{%
\sisetup{detect-all=true}
\begin{tabular}{lS[table-format=2.1]@{\hspace{1.5mm}}>{\fontsize{6}{6}}S[table-format=2.1]S[table-format=2.1]@{\hspace{1.5mm}}>{\fontsize{6}{6}}S[table-format=2.1]S[table-format=2.1]@{\hspace{1.5mm}}>{\fontsize{6}{6}}S[table-format=2.1]S[table-format=2.1]@{\hspace{1.5mm}}>{\fontsize{6}{6}}S[table-format=2.1]S[table-format=2.1]@{\hspace{1.5mm}}>{\fontsize{6}{6}}S[table-format=2.1]S[table-format=2.1]@{\hspace{1.5mm}}>{\fontsize{6}{6}}S[table-format=2.1]S[table-format=2.1]@{\hspace{1.5mm}}>{\fontsize{6}{6}}S[table-format=2.1]}
\toprule
Attr& \multicolumn{2}{c}{\oursnew} & \multicolumn{2}{c}{\nonavos} & \multicolumn{2}{c}{\nosvos} & \multicolumn{2}{c}{\nmsk} & \multicolumn{2}{c}{\nctn} & \multicolumn{2}{c}{\nvpn} & \multicolumn{2}{c}{\nofl} \\
\midrule
LR	&       89.3 &\itshape -3.6	&\bfseries 89.5 &\itshape -5.3	&       80.1 &\itshape 0.1	&       78.9 &\itshape -1.8	&       69.0 &\itshape 3.2	&       57.7 &\itshape 13.5	&       45.7 &\itshape 26.7	\\
SV	&\bfseries 82.9 &\itshape 6.2	&       82.3 &\itshape 5.4	&       74.8 &\itshape 9.1	&       71.8 &\itshape 9.6	&       62.5 &\itshape 14.8	&       58.6 &\itshape 15.3	&       51.0 &\itshape 24.5	\\
FM	&\bfseries 85.2 &\itshape 2.1	&       84.2 &\itshape 1.9	&       77.7 &\itshape 3.9	&       75.0 &\itshape 4.0	&       65.8 &\itshape 8.7	&       57.4 &\itshape 16.1	&       48.7 &\itshape 26.1	\\
CS	&\bfseries 89.8 &\itshape -5.0	&       88.3 &\itshape -4.3	&       80.8 &\itshape -0.9	&       76.8 &\itshape 1.1	&       71.8 &\itshape -0.6	&       68.8 &\itshape -1.5	&       64.2 &\itshape 2.3	\\
DB	&\bfseries 82.8 &\itshape 4.4	&       75.0 &\itshape 12.4	&       75.3 &\itshape 5.8	&       72.5 &\itshape 5.9	&       60.4 &\itshape 13.0	&       42.0 &\itshape 30.4	&       42.8 &\itshape 27.0	\\
MB	&\bfseries 82.8 &\itshape 6.8	&       80.8 &\itshape 8.5	&       74.7 &\itshape 9.9	&       72.1 &\itshape 9.9	&       66.1 &\itshape 9.5	&       62.1 &\itshape 10.4	&       53.6 &\itshape 22.0	\\
OCC	&\bfseries 86.8 &\itshape -0.4	&       84.0 &\itshape 2.1	&       79.8 &\itshape 0.6	&       75.8 &\itshape 2.5	&       70.8 &\itshape 0.8	&       73.2 &\itshape -7.7	&       66.2 &\itshape -0.7	\\
OV	&\bfseries 82.4 &\itshape 5.2	&       80.8 &\itshape 5.9	&       71.1 &\itshape 11.4	&       68.3 &\itshape 11.6	&       63.9 &\itshape 9.3	&       53.8 &\itshape 17.5	&       48.5 &\itshape 21.5	\\
\bottomrule
\end{tabular}}
\vspace{2mm}
\caption{\label{tab:evaltableattr} \textbf{Attribute-based performance ($\J\&\F$)}: Impact of the attributes of the sequences on the results.
For each attribute, results on the sequences with that particular feature and in italics the gain with respect to those on the set of sequences without the attribute. LR stands for low resolution, SV for scale variation, FM for fast motion, CS for camera shake, DB for dynamic background, MB for motion blur, OCC for occlusions, and OV for object out of view.
}
\vspace{-3mm}
\end{table}

\paragraph*{\textbf{Number of training images (parent network)}}
To evaluate how many annotated data are needed to retrain a parent network, Table~\ref{tab:numtrainimages} shows the performance of OSVOS$^S$ when using a subset of the DAVIS 2016 training set. We directly used the output of the CNN, without snapping, for efficiency. We randomly selected a fixed percentage of the annotated frames over all videos of the training set, and evaluated using the Region Similarity ($\J$) metric.
\begin{table}[h]
\centering
\resizebox{0.75\linewidth}{!}{%
\rowcolors{2}{white}{rowblue}
\begin{tabular}{cccccccc}
\toprule
Training data  & 100 		& 200  		& 600 	   & 1000     & 2079  \\ \midrule
Quality ($\mathcal{J}$)  & 82.3  & 84.9   & 85.2   & 85.5   & 85.6   \\
\bottomrule
\end{tabular}}
\vspace{2mm}
\caption{\textbf{Amount of training data}: Region similarity ($\J$) as a function of the number of training images for the parent network of \oursnew{}. Full DAVIS 2016 training set is 2079 training data.}
\label{tab:numtrainimages}
\end{table}
We conclude that by using only \texttildelow200 annotated frames, we are able to reach almost the same performance than when using the full DAVIS 2016 training split. Thus, we therefore do not require full video annotations for the training procedure, that are often expensive to acquire.
Even more, since our method is by definition disregarding temporal information, it is natural that the training data do not require to be temporally coherent.

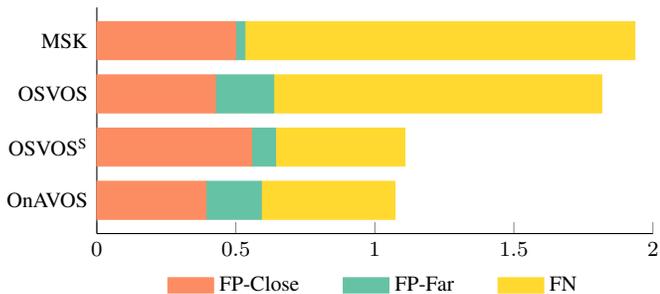
\begin{figure}
\centering
\resizebox{\linewidth}{!}{\begin{tikzpicture}
\begin{axis}[
    xbar stacked,
    legend style={
    legend columns=3,
        at={(xticklabel cs:0.5)},
        anchor=north,
        draw=none
    },
    ytick=data,
    axis y line*=none,
    axis x line*=bottom,
    tick label style={font=\footnotesize},
    legend style={font=\footnotesize},
    label style={font=\footnotesize},
	legend style={/tikz/every even column/.append style={column sep=5mm}},
    width=\linewidth,
    bar width=5mm,
    yticklabels={\nonavos, \oursnew, \ours, \nmsk},
    xmin=0,
    xmax=2,
    area legend,
    y=7mm,
    enlarge y limits={abs=0.625}]
\addplot[Set2-8-2,fill=Set2-8-2] coordinates
{(0.3926,0) (0.5565,1) (0.4264,2) (0.4994,3)};
\addplot[Set2-8-1,fill=Set2-8-1] coordinates
{(0.1989,0) (0.0864,1) (0.2097,2) (0.0332,3)};
\addplot[Set2-8-6,fill=Set2-8-6] coordinates
{(0.4806,0) (0.4651,1) (1.1797,2) (1.4024,3)};
\legend{FP-Close, FP-Far, FN}
\end{axis}  
\end{tikzpicture}}
\vspace{-3mm}
\caption{\textbf{Error analysis of our method}: Errors divided into False Positives (FP-Close and FP-Far) and False Negatives (FN). Values are percentage (\%) of FP-Close, FP-Far or FN pixels in a sequence.}
\label{fig:error_stats}
\end{figure}

\paragraph*{\textbf{Misclassified-Pixels Analysis}}
Figure~\ref{fig:error_stats} shows the error analysis of our method.
We divide the misclassified pixels in three categories:
Close False Positives (FP-Close), Far False Positives (FP-Far) and False Negatives (FN):
(i) FP-Close are those near the contour of the object of interest, so contour inaccuracies,
(ii) FP-Far reveal if the method detects other objects or blobs apart from the object of interest, and
(iii) FN tell us if we miss a part of the object during the sequence.
The measure in the plot is the percentage of pixels in a sequence that fall in a certain category.

The main strength of \oursnew{} compared to \ours{} and \nmsk{} is considerably reducing the number of false negatives.
We believe this is due to \oursnew{}'s ability to \textit{complete} the object of interest when parts that were occluded in the first frame become visible, thanks to the semantic concept of instance.
On the other hand, the output of the instance segmentation network that we are currently using, FCIS~\cite{Dai2016a}, is not very precise on the boundaries of the objects, and even though our conditional classifier is able to recover in part, FP-Close is slightly worse than that of the competition.
On the plus side, since the instance segmentation is an independent input to our algorithm, we will probably directly benefit from better instance segmentation algorithms.

\paragraph*{\textbf{Performance Decay}}
As indicated by the $\J$-Decay and $\F$-Decay values in Table~\ref{tab:evaltable}, \oursnew{} exhibits a better ability than \ours{} and \nmsk{} to maintain performance as frames evolve, and we interpret that this is so thanks to the injected semantic prior.
The performance decay is similar to that of \nonavos{}, even though it performs a costly iterative algorithm which fine-tunes the result to various frames of the sequence. Our method, on the other hand, uses the information of the first frame only, and keeps the quality throughout the sequence.

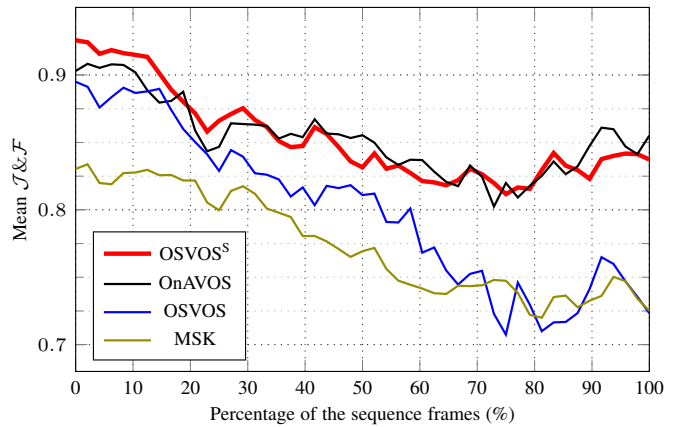
\begin{figure}
\resizebox{\linewidth}{!}{%
\begin{tikzpicture}[/pgfplots/width=1\linewidth, /pgfplots/height=0.7\linewidth]
    \begin{axis}[ymin=0.68,ymax=0.95,xmin=0,xmax=1,enlargelimits=false,
        xlabel=Percentage of the sequence frames (\%),
        ylabel=Mean $\mathcal{J}\&\mathcal{F}$,
        font=\scriptsize, grid=both,
        legend pos= south west,
        grid style=dotted,
        axis equal image=false,
        ytick={0.1,0.2,...,0.9},
        xtick={0,0.1,...,1.1},
        xticklabels={0,10,20,30,40,50,60,70,80,90,100},
        minor ytick={0.1,0.125,...,0.9},
        major grid style={white!20!black},
        minor grid style={white!70!black},
        xlabel shift={-2pt},
        ylabel shift={-3pt}] 
                 \addplot+[red,solid,mark=none, ultra thick] table[x=time,y=OSVOS-S] {data/performance_drop_curve.txt};
                 \addlegendentry{\oursnew}
         \addplot+[black,solid,mark=none, thick] table[x=time,y=OnAVOS] {data/performance_drop_curve.txt};
                  \addlegendentry{\nonavos}
         \addplot+[blue,solid,mark=none, thick] table[x=time,y=OSVOS] {data/performance_drop_curve.txt};
         \addlegendentry{\ours}
         \addplot+[olive,solid,mark=none, thick] table[x=time,y=MaskTrack_Flow_CRF] {data/performance_drop_curve.txt};
         \addlegendentry{\nmsk}

	 \end{axis}
   \end{tikzpicture}}
   \caption{\textbf{Decay of the quality with time}: Performance of various methods with respect to the \textit{time} axis.}
   \label{fig:decay}
\end{figure}

To further highlight this result and analyze it more in detail, 
Figure~\ref{fig:decay} shows the evolution of $\J$ as the sequence advances, to examine how the performance drops over time.
Since the videos in DAVIS 2016 are of different length, we normalize them to $[0,100]$ as a percentage of the sequence length.
We then compute the mean $\J$ curve among all video sequences.
As it can be seen from Figure~\ref{fig:decay}, our method is significantly more stable in terms of performance drop compared to \ours{} and \nmsk{}, and has a similar curve than \nonavos{}.

We also report the lowest point of the curve which indicates the worst performance across the video.
Based on this metrics, \oursnew{} is at $82.0$, while for semantic-blind methods, the numbers are $81.0$, $73.7$, and $69.8$.

The results therefore confirm that the semantic prior we introduce can mitigate the performance drop caused by appearance change, while maintaining high fidelity in details.
The semantic information is particularly helpful in the later stage of videos where dramatic appearance changes with respect to the first frame are more probable.

\paragraph*{\textbf{Speed}}
The computational efficiency of video object segmentation is crucial for the algorithms to be usable in practice.
\oursnew{} can adapt to different timing requirements, providing progressively better results 
the more time we can afford, by letting the fine-tuning algorithm at test time do more or fewer iterations.
As introduced before, \oursnew{}'s time can be divided into the fine-tuning time plus the time to process each frame independently.

To compare to other techniques, we will evaluate the mean computing time per frame: 
fine-tuning time (done once per sequence) averaged over the length of that sequence, plus the forward pass on each frame.

\begin{figure}[t]
\centering
\resizebox{\linewidth}{!}{\begin{tikzpicture}[/pgfplots/width=0.85\linewidth, /pgfplots/height=0.68\linewidth, /pgfplots/legend pos=south east]
    \begin{axis}[ymin=0.565,ymax=0.87,xmin=9e-2,xmax=60,enlargelimits=false,
        xlabel=Time per frame ($s$),
        ylabel=Performance ($\mathcal{J}\&\mathcal{F}$),
		font=\scriptsize,
        grid=both,
		grid style=dotted,
        xlabel shift={-2pt},
        ylabel shift={-5pt},
        xmode=log,
        legend columns=1,
        legend style={/tikz/every even column/.append style={column sep=3mm}},
        minor ytick={0,0.025,...,1.1},
        ytick={0,0.1,...,1.1},
		yticklabels={0,.1,.2,.3,.4,.5,.6,.7,.8,.9,1},
	    xticklabels={0,.1,1,10,100},
        legend pos= outer north east
        ]

        \addplot+[smooth,black,mark=none, line width=1] table[x=time,y=j_val] {data/quality_vs_time_inst_ucm.txt};
        \addlegendentry{\oursnew{}}
        \label{fig:qual_vs_time:osvoss}
        \addplot+[smooth,forget plot,black,mark=none, line width=0.5, dashed] table[x=time,y=j_val] {data/quality_vs_time_inst_ucm_dashed.txt};

        \addplot+[smooth,red,mark=none, line width=1] table[x=time,y=j_val] {data/quality_vs_time_inst.txt};
        \addlegendentry{No Snapping}
      \label{fig:qual_vs_time:nosnapping}
        \addplot+[smooth,forget plot,red,mark=none, line width=0.5, dashed] table[x=time,y=j_val] {data/quality_vs_time_inst_dashed.txt};
         \addplot+[smooth,forget plot,red,mark=none, line width=0.5, dashed] table[x=time,y=j_val] {data/quality_vs_time_inst_dashed2.txt};

		\addplot+[smooth,blue,mark=none, line width=1] table[x=time,y=j_val] {data/quality_vs_time_base.txt};
        \addlegendentry{No Semantic}
        \label{fig:qual_vs_time:nosemantic}
        
        \addplot+[smooth,forget plot,blue,mark=none, line width=0.5, dashed] table[x=time,y=j_val] {data/quality_vs_time_base_dashed.txt};

        \addplot+[forget plot,black,mark=square*, mark size=1.1,only marks, mark options={fill=black}] coordinates{(0.958467,0.858)};
        \label{fig:qual_vs_time:osvosspre}

        \addplot+[forget plot,red,mark=square*, mark size=1.1,only marks, mark options={fill=red}] coordinates{(0.343834,0.855)};
        \label{fig:qual_vs_time:nosnappingpre}
        
        \addplot+[forget plot,blue,mark=square*,solid,mark size=1.1,only marks, mark options={fill=blue}] coordinates{(0.129834,0.77352)};
        \label{fig:qual_vs_time:nosemanticspre}

        \addplot[olive,mark=asterisk, mark size=1.9,only marks, line width=0.75] coordinates{(13,0.8551)};
        \addlegendentry{\nonavos~\cite{\onavos}}
		\label{fig:qual_vs_time:onavos}
		
        \addplot[orange,mark=+,only marks,line width=0.75] coordinates{(12,0.775)};
        \addlegendentry{\nmsk~\cite{\msk}}
        \label{fig:qual_vs_time:msk}
        
        \addplot[black,mark=asterisk,only marks,line width=0.75] coordinates{(0.63,0.678)};
        \addlegendentry{\nvpn~\cite{\vpn}}
        \label{fig:qual_vs_time:vpn}
               
        \addplot[blue,mark=Mercedes star,only marks,line width=0.75] coordinates{(29.95,0.714)};
        \addlegendentry{\nctn~\cite{\ctn}}
        \label{fig:qual_vs_time:ctn}
		
        \addplot[cyan,mark=asterisk, mark size=2.2,only marks, line width=0.75] coordinates{(42,0.68)};
        \addlegendentry{\nofl~\cite{\ofl}}
		\label{fig:qual_vs_time:ofl}
		
       	\addplot[green,mark=Mercedes star,only marks,line width=0.75] coordinates{(0.37,0.594)};
        \addlegendentry{\nbvs~\cite{\bvs}}
        \label{fig:qual_vs_time:bvs}
           
    \end{axis}
\end{tikzpicture}}
\vspace{-7mm}
   \caption{\textbf{Quality versus timing}: $\mathcal{J}\&\mathcal{F}$ with respect to the processing time per frame.}
   \label{fig:qual_vs_time}
   \vspace{-2mm}
\end{figure}
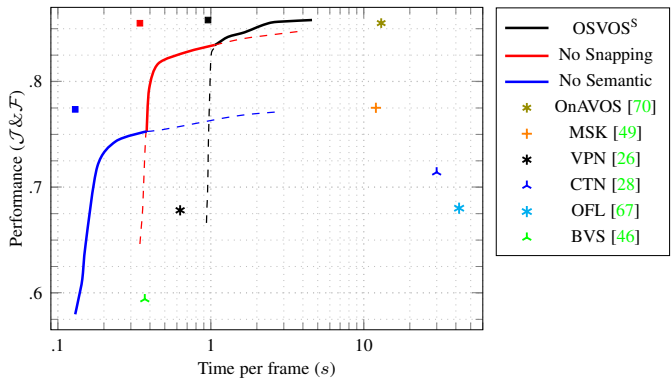

Figure~\ref{fig:qual_vs_time} shows the quality of the result with respect to the time it takes to process each 480p frame. The computation time for our method has been obtained using an NVidia Titan X GPU and for other methods the timing reported in their publications has been used.
Our techniques are represented by curves: \oursnew{} (\ref{fig:qual_vs_time:osvoss}), without boundary snapping (\ref{fig:qual_vs_time:nosnapping}), and without semantics (\ref{fig:qual_vs_time:nosemantic}), which show the gain in quality with respect to the fine-tuning time.
The best results come at the price of adding the semantics or the snapping cost, so depending on the needed speed, one of the three modes can be chosen.
Dashed lines represent the regimes of each technique that are not in the Pareto front, \ie where it is better to choose another mode within our techniques (faster for the same quality or best quality for the same speed).

Since \oursnew{} processes frames independently, one could also perform the fine-tuning offline, by training on a picture of the object to be segmented beforehand (\eg take a picture of a sports player before a match).
In this scenario, \oursnew{} can process each frame by one forward pass of the CNN (\ref{fig:qual_vs_time:osvosspre} $|$ \ref{fig:qual_vs_time:nosnappingpre} $|$ \ref{fig:qual_vs_time:nosemanticspre}), and so be considerably fast.

Compared to other techniques, our techniques are faster and/or more accurate at all regimes, from fast modes: $75.1$ versus $59.4$ of \nbvs{} (\ref{fig:qual_vs_time:bvs}) at 300 miliseconds, to high-quality regimes: same performance than \nonavos{} (\ref{fig:qual_vs_time:onavos}) but an order of magnitude faster (2.5 versus 12 seconds). 
The trade-off between performance and speed in video object segmentation has been largely ignored (or purposely hidden) in the literature although we believe it is of critical importance, and so we encourage future research to evaluate their methods in this performance-vs-speed plane.

\paragraph*{\textbf{Comparison to State of the Art in Youtube-Objects}}
For completeness, we also do experiments on Youtube-objects~\cite{Prest2012,Jain2014}, without changing any parameter from our algorithm nor retraining the parent network.
Table~\ref{tab:youtube} shows the results of the quantitative evaluation against the rest of techniques.
\oursnew{} obtains the best results overall, being two points better than the runner up; and having the best results in eight out of ten categories.
These experiments show the robustness and generality of our approach even to domain (dataset) shifts.
 
\begin{table}[t]
\centering
\rowcolors{2}{white}{rowblue}
\resizebox{1\linewidth}{!}{%
\begin{tabular}{lcccccc}
\toprule
Category  & \oursnew  & \nonavos   & \nosvos    & \nmsk      & \nofl      & \nbvs      \\
\midrule
Aeroplane & \bf\ 90.4 &    \ 87.7 &    \ 88.2 &    \ 86.0 &    \ 89.9 &    \ 86.8 \\
Bird      & \bf\ 87.0 &    \ 85.7 &    \ 85.7 &    \ 85.6 &    \ 84.2 &    \ 80.9 \\
Boat      & \bf\ 83.6 &    \ 78.5 &    \ 77.5 &    \ 78.8 &    \ 74.0 &    \ 65.1 \\
Car       & \bf\ 87.9 &    \ 86.1 &    \ 79.6 &    \ 78.8 &    \ 80.9 &    \ 68.3 \\
Cat       & \bf\ 80.7 &    \ 80.5 &    \ 70.8 &    \ 70.1 &    \ 68.3 &    \ 55.8 \\
Cow       &    \ 79.3 &    \ 77.9 &    \ 77.8 &    \ 77.7 & \bf\ 79.8 &    \ 69.9 \\
Dog       & \bf\ 82.5 &    \ 80.8 &    \ 81.3 &    \ 79.2 &    \ 76.6 &    \ 68.0 \\
Horse     & \bf\ 73.9 &    \ 72.1 &    \ 72.8 &    \ 71.7 &    \ 72.6 &    \ 58.9 \\
Motorbike & \bf\ 79.3 &    \ 72.0 &    \ 73.5 &    \ 65.6 &    \ 73.7 &    \ 60.5 \\
Train     & \bf\ 87.1 &    \ 84.0 &    \ 75.7 &    \ 83.5 &    \ 76.3 &    \ 65.2 \\
\midrule
Mean      & \bf\ 83.2 &    \ 80.5 &    \ 78.3 &    \ 77.7 &    \ 77.6 &    \ 67.9 \\
\bottomrule
\end{tabular}}
\vspace{2mm}
\caption{\textbf{Youtube-Objects evaluation}: Per-category and overall mean intersection over union ($\J$).}
\label{tab:youtube}
\end{table}

\begin{figure*}
\centering
\resizebox{\textwidth}{!}{%
	  \setlength{\fboxsep}{0pt}
      \rotatebox{90}{\hspace{4.5mm}Drift-Chicane\vphantom{p}}
      \fbox{\includegraphics[width=0.3\textwidth]{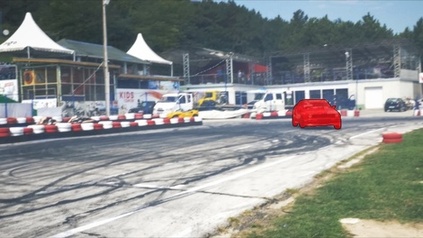}}   
      \fbox{\includegraphics[width=0.3\textwidth]{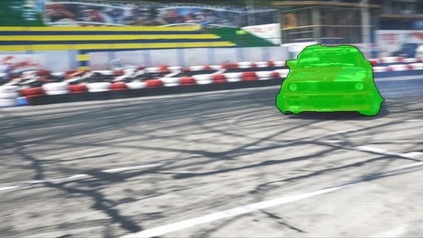}}
      \fbox{\includegraphics[width=0.3\textwidth]{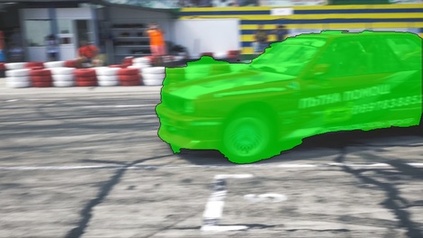}}
      \fbox{\includegraphics[width=0.3\textwidth]{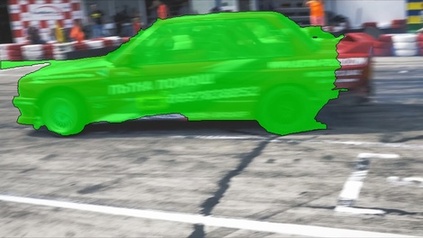}}
      \fbox{\includegraphics[width=0.3\textwidth]{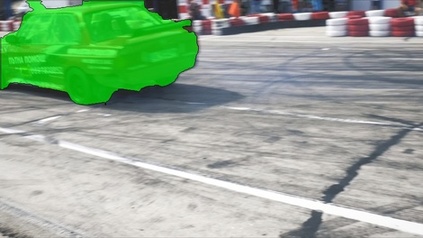}}
      }\\[1mm]
\resizebox{\textwidth}{!}{%
	  \setlength{\fboxsep}{0pt}
      \rotatebox{90}{\hspace{2.5mm}Motocross-Jump}
      \fbox{\includegraphics[width=0.3\textwidth]{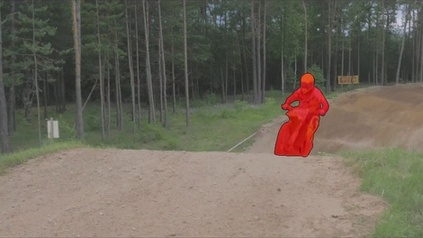}}
      \fbox{\includegraphics[width=0.3\textwidth]{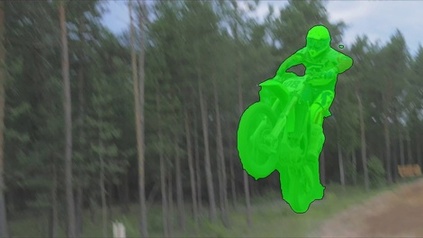}}
      \fbox{\includegraphics[width=0.3\textwidth]{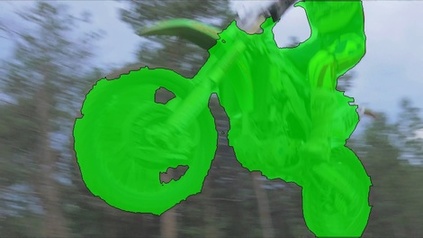}}
      \fbox{\includegraphics[width=0.3\textwidth]{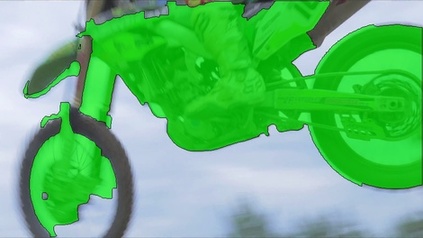}}
      \fbox{\includegraphics[width=0.3\textwidth]{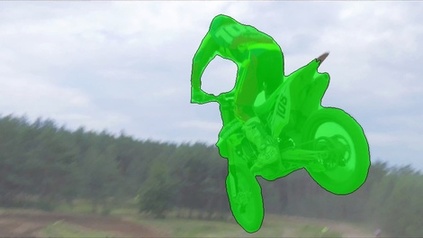}}
      }\\[1mm]
\resizebox{\textwidth}{!}{%
	  \setlength{\fboxsep}{0pt}
      \rotatebox{90}{\hspace{7.5mm}Kite-Surf\vphantom{p}}
      \fbox{\includegraphics[width=0.3\textwidth]{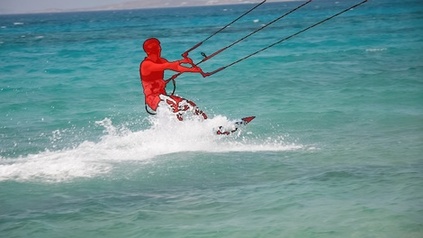}}
      \fbox{\includegraphics[width=0.3\textwidth]{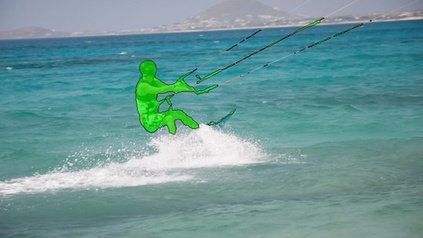}}
      \fbox{\includegraphics[width=0.3\textwidth]{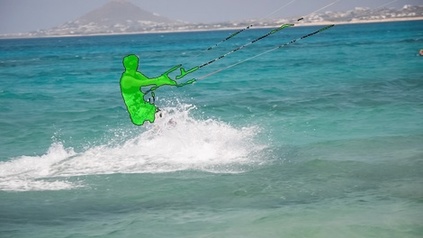}}
      \fbox{\includegraphics[width=0.3\textwidth]{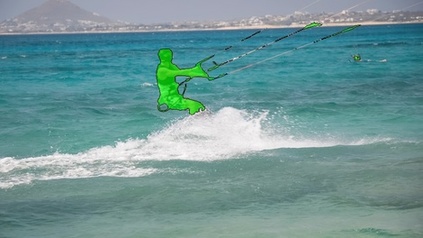}}
      \fbox{\includegraphics[width=0.3\textwidth]{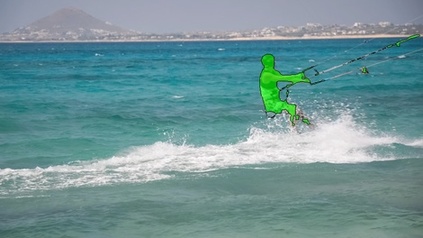}}
      }\\[1mm]
\resizebox{\textwidth}{!}{%
	  \setlength{\fboxsep}{0pt}
      \rotatebox{90}{\hspace{9mm}Libby\vphantom{p}}
      \fbox{\includegraphics[width=0.3\textwidth]{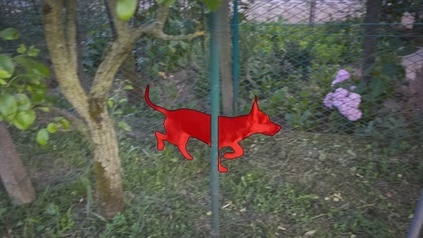}}
      \fbox{\includegraphics[width=0.3\textwidth]{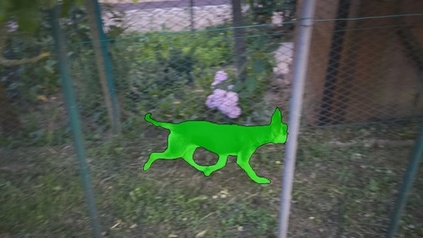}}
      \fbox{\includegraphics[width=0.3\textwidth]{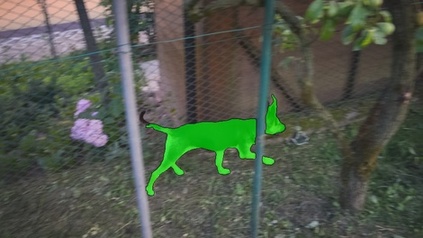}}
      \fbox{\includegraphics[width=0.3\textwidth]{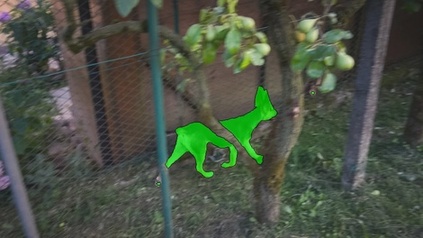}}
      \fbox{\includegraphics[width=0.3\textwidth]{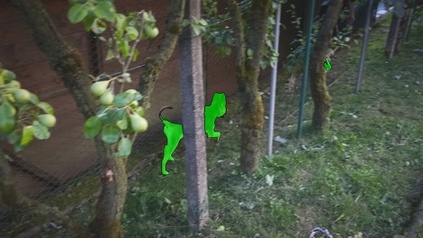}}
      }\\[1mm]
\resizebox{\textwidth}{!}{%
	  \setlength{\fboxsep}{0pt}
      \rotatebox{90}{\hspace{2.5mm}Horsejump-High\vphantom{p}}
      \fbox{\includegraphics[width=0.3\textwidth]{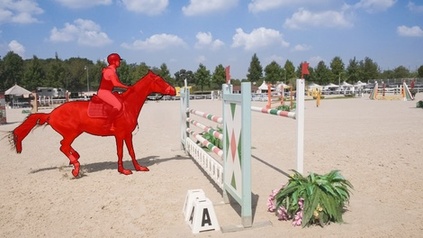}}
      \fbox{\includegraphics[width=0.3\textwidth]{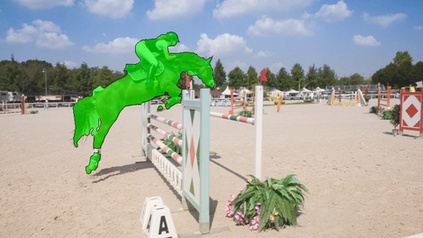}}
      \fbox{\includegraphics[width=0.3\textwidth]{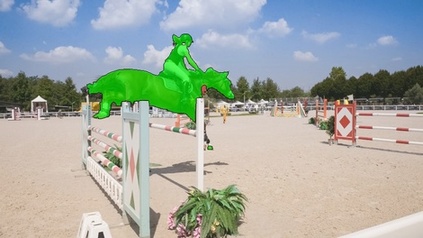}}
      \fbox{\includegraphics[width=0.3\textwidth]{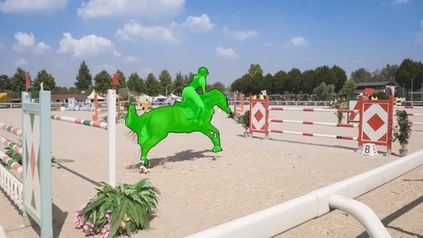}}
      \fbox{\includegraphics[width=0.3\textwidth]{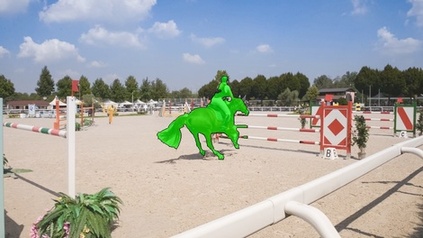}}
      }\\[1mm]   
\resizebox{\textwidth}{!}{%
	  \setlength{\fboxsep}{0pt}
      \rotatebox{90}{\hspace{3.5mm}Car-Roundabout\vphantom{p}}
      \fbox{\includegraphics[width=0.3\textwidth]{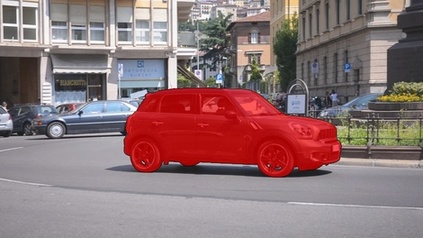}}
      \fbox{\includegraphics[width=0.3\textwidth]{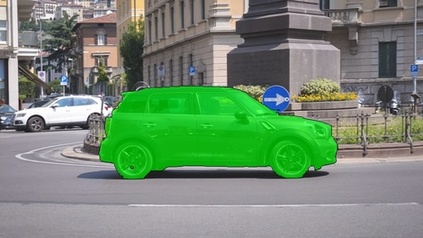}}
      \fbox{\includegraphics[width=0.3\textwidth]{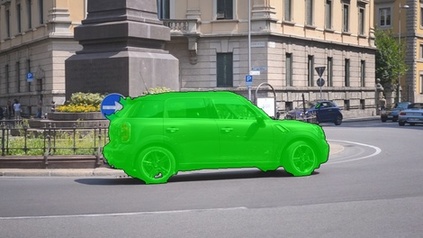}}
      \fbox{\includegraphics[width=0.3\textwidth]{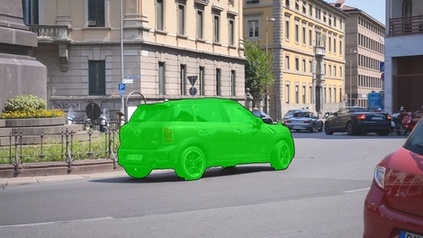}}
      \fbox{\includegraphics[width=0.3\textwidth]{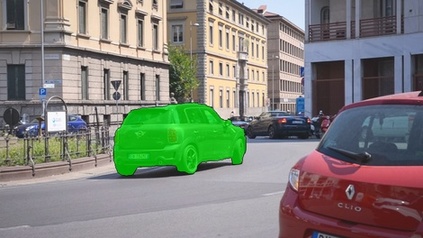}}
      }\\[1mm]
\resizebox{\textwidth}{!}{%
	  \setlength{\fboxsep}{0pt}
      \rotatebox{90}{\hspace{10.5mm}Camel\vphantom{p}}
      \fbox{\includegraphics[width=0.3\textwidth]{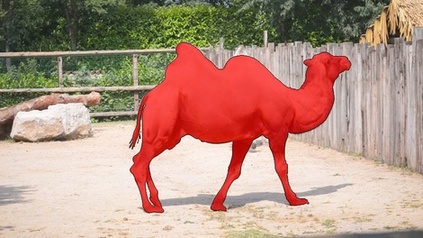}}
      \fbox{\includegraphics[width=0.3\textwidth]{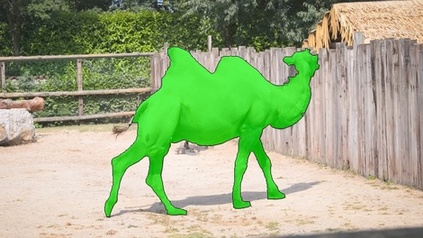}}
      \fbox{\includegraphics[width=0.3\textwidth]{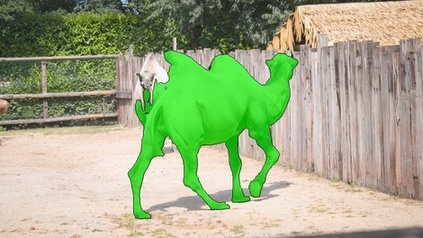}}
      \fbox{\includegraphics[width=0.3\textwidth]{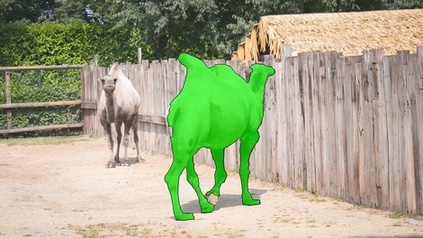}}
      \fbox{\includegraphics[width=0.3\textwidth]{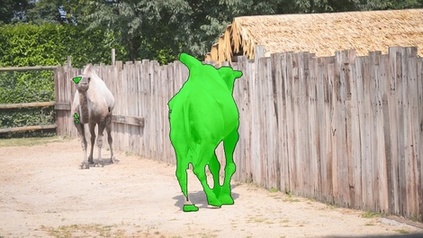}}
      }\\[1mm]
\resizebox{\textwidth}{!}{%
	  \setlength{\fboxsep}{0pt}
      \rotatebox{90}{\hspace{8.5mm}Soapbox\vphantom{p}}
      \fbox{\includegraphics[width=0.3\textwidth]{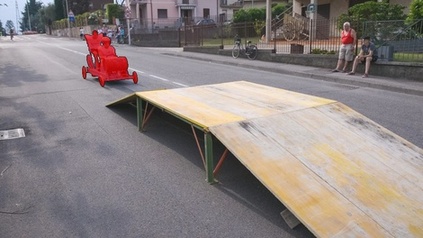}}
      \fbox{\includegraphics[width=0.3\textwidth]{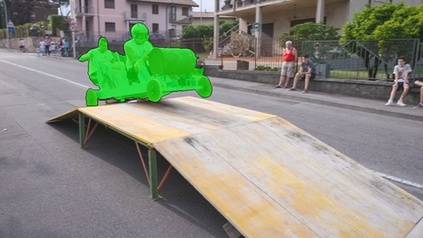}}
      \fbox{\includegraphics[width=0.3\textwidth]{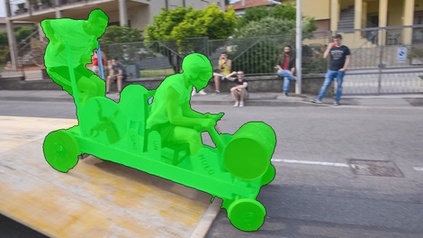}}
      \fbox{\includegraphics[width=0.3\textwidth]{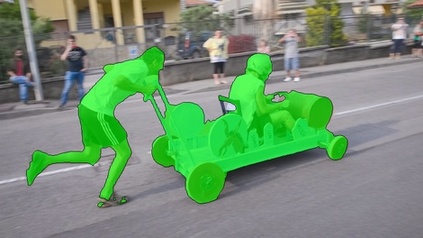}}
      \fbox{\includegraphics[width=0.3\textwidth]{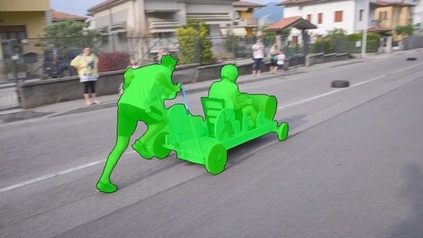}}
      }\\[1mm]   
\resizebox{\textwidth}{!}{%
	  \setlength{\fboxsep}{0pt}
      \rotatebox{90}{\hspace{6.5mm}Breakdance\vphantom{p}}
      \fbox{\includegraphics[width=0.3\textwidth]{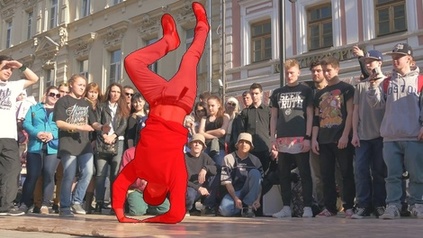}}
      \fbox{\includegraphics[width=0.3\textwidth]{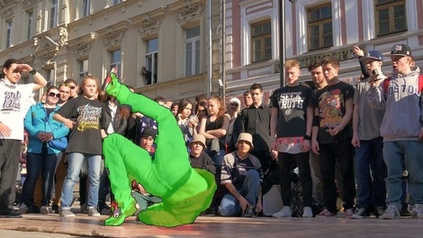}}
      \fbox{\includegraphics[width=0.3\textwidth]{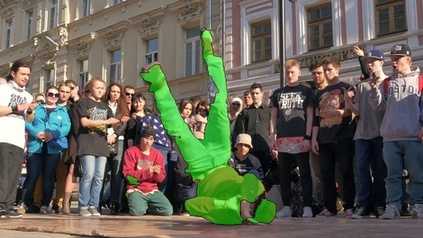}}
      \fbox{\includegraphics[width=0.3\textwidth]{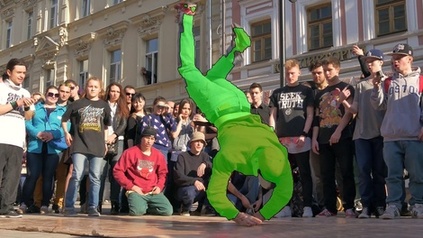}}
      \fbox{\includegraphics[width=0.3\textwidth]{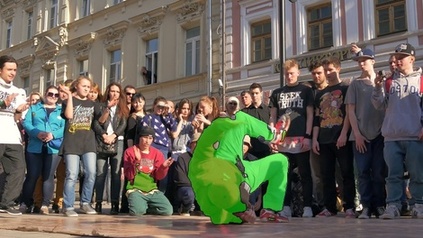}}
      }\\[1mm]   
\resizebox{\textwidth}{!}{%
	  \setlength{\fboxsep}{0pt}
      \rotatebox{90}{\hspace{6.5mm}Bmx-Trees\vphantom{p}}
      \fbox{\includegraphics[width=0.3\textwidth]{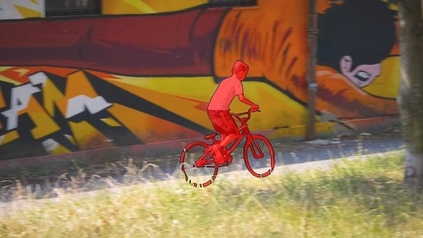}}
      \fbox{\includegraphics[width=0.3\textwidth]{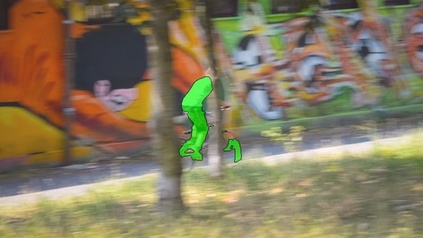}}
      \fbox{\includegraphics[width=0.3\textwidth]{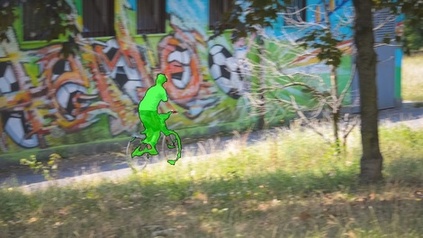}}
      \fbox{\includegraphics[width=0.3\textwidth]{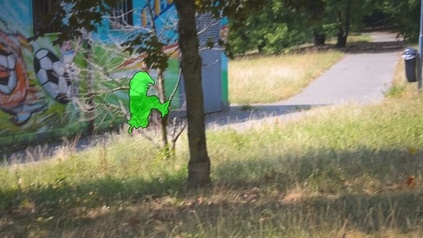}}
      \fbox{\includegraphics[width=0.3\textwidth]{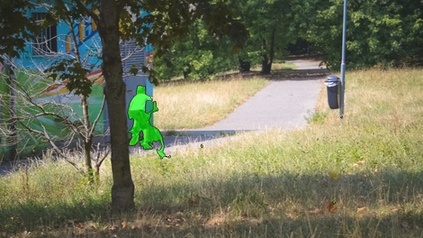}}
      }
\caption{\textbf{Qualitative results on DAVIS 2016}: \oursnew{} results on a variety of representatives sequences. The input to our algorithm is the ground truth of the first frame (red). Outputs of all frames (green) are produced independent of each other.}
\label{fig:qualitative}
\vspace{-2mm}
\end{figure*}

\begin{figure*}
\centering
\resizebox{\textwidth}{!}{%
	  \setlength{\fboxsep}{0pt}
      \rotatebox{90}{\hspace{11.5mm}Car\vphantom{p}}
      \fbox{\includegraphics[width=0.3\textwidth]{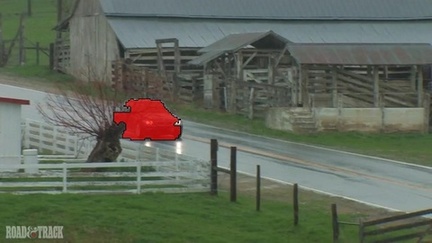}}   
      \fbox{\includegraphics[width=0.3\textwidth]{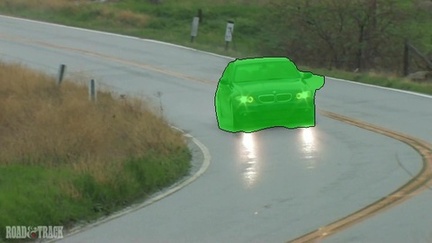}}
      \fbox{\includegraphics[width=0.3\textwidth]{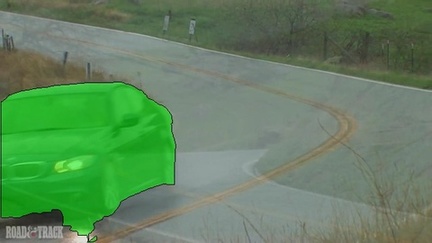}}
      \fbox{\includegraphics[width=0.3\textwidth]{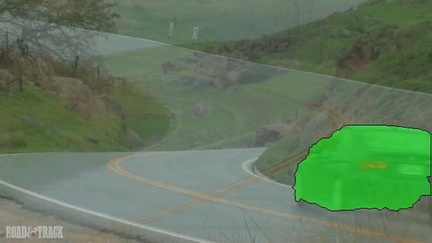}}
      \fbox{\includegraphics[width=0.3\textwidth]{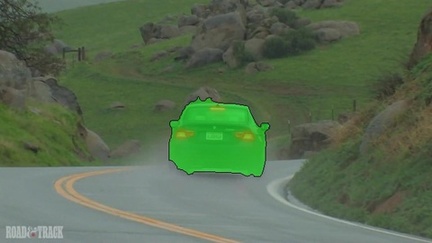}}
      }\\[1mm]
\resizebox{\textwidth}{!}{%
	  \setlength{\fboxsep}{0pt}
      \rotatebox{90}{\hspace{11.5mm}Cat\vphantom{p}}
      \fbox{\includegraphics[width=0.3\textwidth]{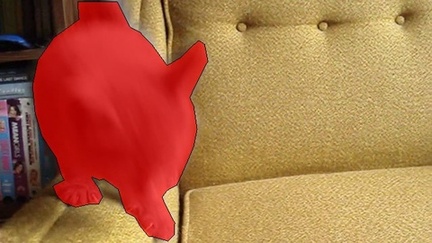}}   
      \fbox{\includegraphics[width=0.3\textwidth]{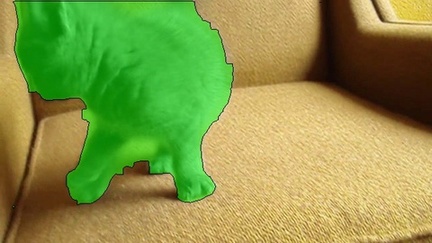}}
      \fbox{\includegraphics[width=0.3\textwidth]{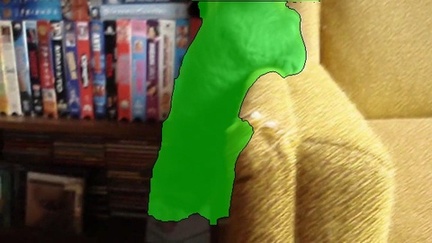}}
      \fbox{\includegraphics[width=0.3\textwidth]{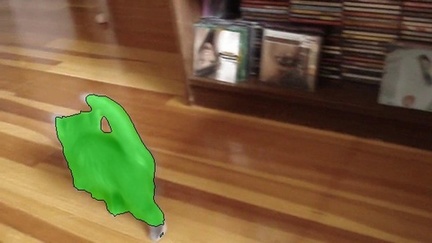}}
      \fbox{\includegraphics[width=0.3\textwidth]{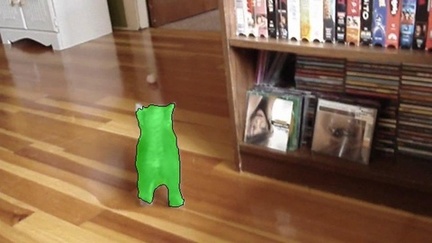}}
      }\\[1mm]
\resizebox{\textwidth}{!}{%
	  \setlength{\fboxsep}{0pt}
      \rotatebox{90}{\hspace{8.5mm}Aeroplane\vphantom{p}}
      \fbox{\includegraphics[width=0.3\textwidth]{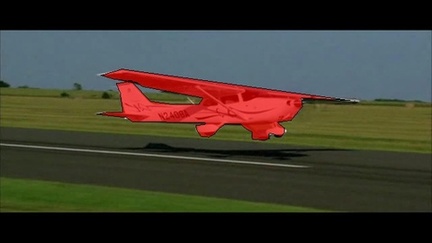}}   
      \fbox{\includegraphics[width=0.3\textwidth]{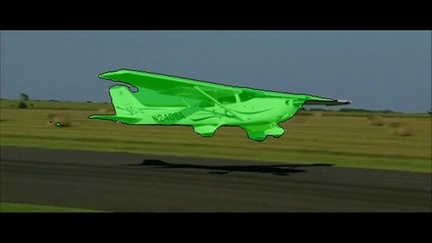}}
      \fbox{\includegraphics[width=0.3\textwidth]{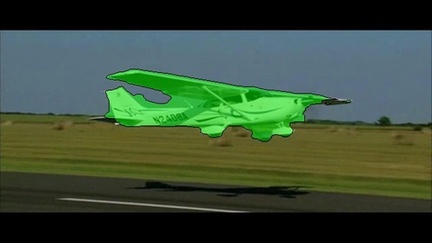}}
      \fbox{\includegraphics[width=0.3\textwidth]{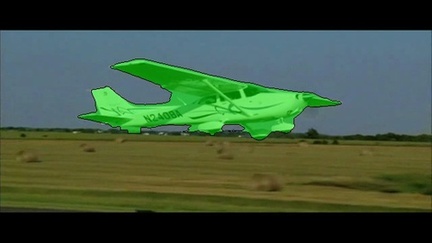}}
      \fbox{\includegraphics[width=0.3\textwidth]{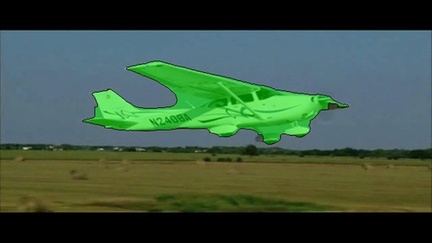}}
      }\\[1mm]
\resizebox{\textwidth}{!}{%
	  \setlength{\fboxsep}{0pt}
      \rotatebox{90}{\hspace{14.5mm}Cow\vphantom{p}}
      \fbox{\includegraphics[width=0.3\textwidth]{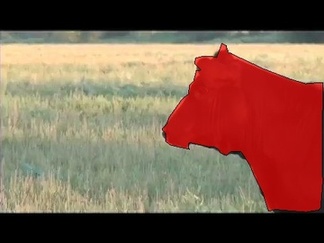}}   
      \fbox{\includegraphics[width=0.3\textwidth]{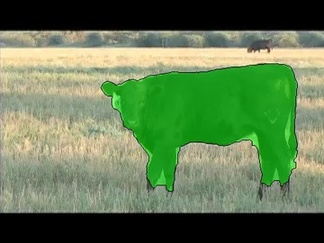}}
      \fbox{\includegraphics[width=0.3\textwidth]{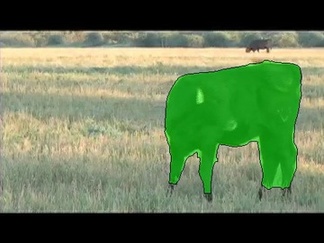}}
      \fbox{\includegraphics[width=0.3\textwidth]{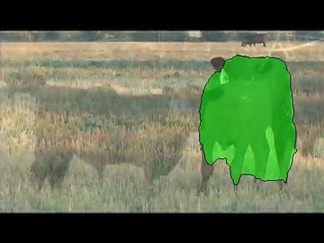}}
      \fbox{\includegraphics[width=0.3\textwidth]{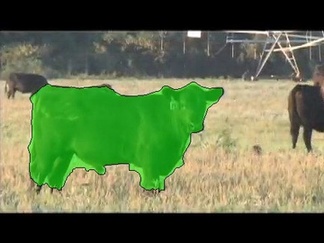}}
      }\\[1mm]
\resizebox{\textwidth}{!}{%
	  \setlength{\fboxsep}{0pt}
      \rotatebox{90}{\hspace{11.5mm}Boat\vphantom{p}}
      \fbox{\includegraphics[width=0.3\textwidth]{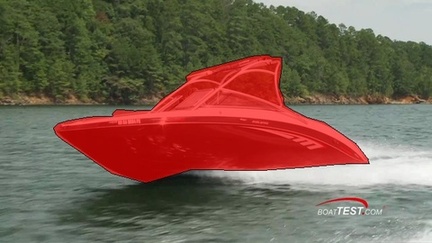}}   
      \fbox{\includegraphics[width=0.3\textwidth]{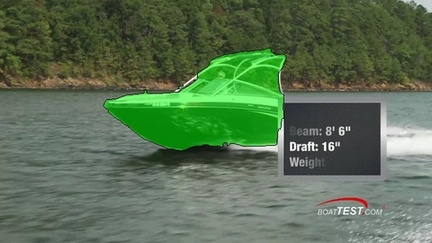}}
      \fbox{\includegraphics[width=0.3\textwidth]{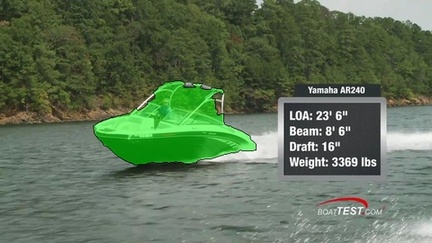}}
      \fbox{\includegraphics[width=0.3\textwidth]{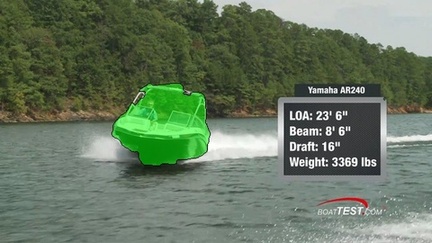}}
      \fbox{\includegraphics[width=0.3\textwidth]{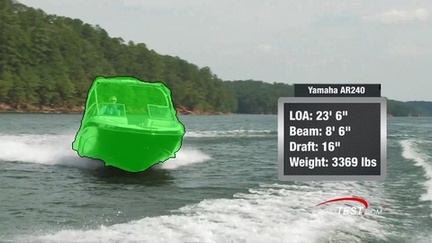}}
      }\\[1mm]
\resizebox{\textwidth}{!}{%
	  \setlength{\fboxsep}{0pt}
      \rotatebox{90}{\hspace{8.5mm}Motorbike\vphantom{p}}
      \fbox{\includegraphics[width=0.3\textwidth]{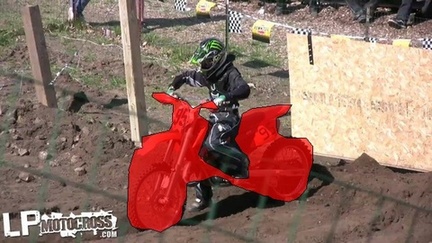}}   
      \fbox{\includegraphics[width=0.3\textwidth]{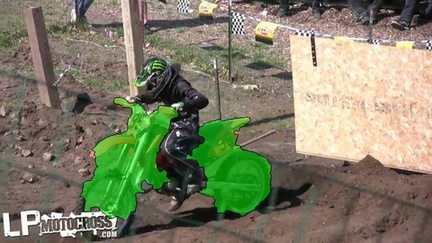}}
      \fbox{\includegraphics[width=0.3\textwidth]{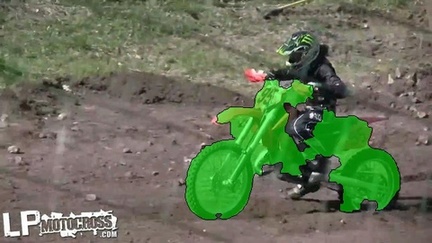}}
      \fbox{\includegraphics[width=0.3\textwidth]{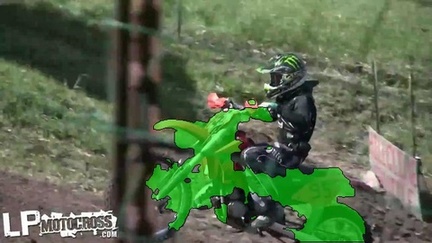}}
      \fbox{\includegraphics[width=0.3\textwidth]{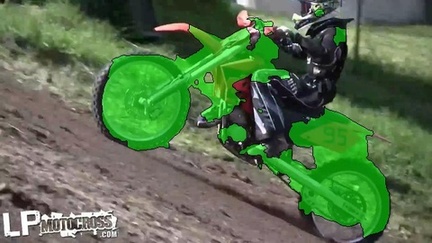}}
      }
\caption{\textbf{Qualitative results on Youtube-Objects}: \oursnew{} results on a variety of representatives sequences. The input to our algorithm is the ground truth of the first frame (red). Outputs of all frames (green) are produced independent of each other.}
\label{fig:qualitative_youtube}
\vspace{-2mm}
\end{figure*}

\paragraph*{\textbf{Multi-object video segmentation in DAVIS 2017}}
We test \oursnew{} in the more challenging DAVIS 2017 dataset where multiple objects have to be segmented in the same video sequence. We apply our method as is, precessing every object in a sequence independently. Table~\ref{tab:davis2017} illustrates the results obtained in the test-dev set of the dataset, compared to the top-performing methods of the DAVIS challenge, and to our direct competitor (OnAVOS).  Even though our method is not specifically designed to handle multiple object instances, we achieve competitive results (comparable to the third entry), and we outperform \nonavos. Our method falls behind the two first entries as it is not optimized to segment multiple objects, and is uses a single model, without the bells and whistles that naturally come with challenge submissions.

\begin{table}[t]
	\centering
	\rowcolors{2}{white}{rowblue}
	\resizebox{.6\linewidth}{!}{%
		\begin{tabular}{lc}
		\toprule
			Method  			  & Test-Dev $\mathcal{J} \& \mathcal{F}$  \\
		\midrule
			Apata~\cite{DAVIS2017-2nd}           	  & \bf  66.6 \\
			Lixx~\cite{DAVIS2017-1st}      			  & \ 66.1  \\
			Wangzhe~\cite{DAVIS2017-3rd}           	  & \ 57.7  \\
			Lalafine123~\cite{DAVIS2017-4th}           & \ 57.4   \\
			Voiglaender~\cite{DAVIS2017-5th}$\qquad$          & \ 56.5  \\
		\midrule
			\nonavos~\cite{Voigtlaender2017}           & \ 52.8 \\
			\oursnew	& \ 57.5 \\
		\bottomrule
	\end{tabular}}
	\vspace{2mm}
	\caption{\textbf{DAVIS 2017 evaluation}: Performance of \oursnew{} compared to the DAVIS 2017 challenge winners, on the test-dev set. Our single model achieves competitive results.}
	\label{tab:davis2017}
\end{table}

\paragraph*{\textbf{Instance segmentation quality}}
In this section we analyze the influence of the quality of the instance segmentation method in our final result.
To this end, we use three different methods, \ie MNC~\cite{dai2016instance}, FCIS~\cite{Li+17}, and Mask-RCNN~\cite{He+17}. Developments to the field over the last two years have lead to competitive results on COCO~\cite{Lin2014} test-dev, with resulting Average Precision (AP) varying from 24.6\% for MNC, to 33.6\% and 37.1\% for FCIS and Mask-RCNN, respectively. Table~\ref{tab:instance_seg} shows the performance gains obtained by using a different instance segmentation method within the \oursnew{} pipeline in three different datasets.
Results suggest that our method is able to incorporate improved instance segmentation results, and directly translates them into more accurate results for video object segmentation. The improvements are particularly large for DAVIS 2017, where there is still room for improvement.

\begin{table}[t]
\centering
\rowcolors{4}{white}{rowblue}
\resizebox{0.9\linewidth}{!}{%
\begin{tabular}{lcccc}
\toprule
        			&  \multicolumn{3}{c}{\oursnew}     & \\
\cmidrule(lr){2-4}			
Dataset  			& \it Mask-RCNN  & \it FCIS   & \it MNC   & \ours \\
\midrule
DAVIS 2016           & \bf\ 86.5 & \   86.0  &    \ 83.5 &    \ 80.2 \\
Youtube-Objects      & \bf\ 83.2 & \   82.5  &    \ 80.8 &    \ 78.3  \\
DAVIS 2017           & \bf\ 57.5 & \   53.7 &    \ 51.5 &    \ 48.7   \\

\bottomrule
\end{tabular}}
\vspace{2mm}
\caption{\textbf{Performance vs. instance segmentation quality}: Evaluation with respect to the instance segmentation algorithm.}
\label{tab:instance_seg}
\end{table}

\paragraph*{\textbf{Qualitative Results}}
Figure~\ref{fig:qualitative} and Figure~\ref{fig:qualitative_youtube} show some qualitative results of \oursnew{} in DAVIS 2016 and Youtube-Objects, respectively.
The first column shows the ground-truth mask used as input to our algorithm (in red).
The rest of the columns show our segmented results in the following frames.
These visual results qualitatively corroborate the robustness of our approach to occlusions, dynamic background, change of appearance, etc.

\paragraph*{\textbf{Limitations of \oursnew}}
Both \ours{} and \oursnew{} are very practical for applications due to their accuracy, and their frame-independent design which comes with increased speed compared to competing methods. Limitations of \ours{} mainly regard appearance of objects, such as similar objects, dynamic changes in appearance and viewpoint, and are successfully tackled by introducing the coarse instance segmentation input in \oursnew{}. False positives can be successfully tackled by introducing optical flow models~\cite{DAVIS2017-2nd}, whereas~\cite{DAVIS2017-1st} handle false negatives by introducing a re-identification module, with the cost of extra processing time. Limitations regarding out-of-vocabulary instances are handled well by our method, however, that may not transfer to other domains with uncommon objects or parts of objects.

\section{Conclusions}
This paper presents \longoursnew{} (\oursnew{}), a semi-supervised video object segmentation technique that
processes each frame independently and thus ignores the temporal information and redundancy of a video sequence.
This has the inherent advantage of being robust to object occlusions, lost frames, etc, while keeping execution speed low.

\oursnew{} shows state-of-the-art results in both DAVIS 2016 and Youtube-Objects at the whole range of operating speeds. It is significantly faster and/or better performing than the competition: $75.1$ versus $59.4$ at 300 miliseconds per frame, or 4.5 versus 12 seconds at the best performance (86.5 vs 85.5).

To do so, we build a powerful appearance model of the object from a single segmented frame.
In contrast to most deep learning approaches, that often require a huge amount of training data in order to solve a specific problem, and in line with humans, that can solve similar challenges with only a single training example; we demonstrate that \oursnew{} can reproduce this capacity of one-shot learning in a machine: Based on a parent network architecture pre-trained on a generic video segmentation dataset, we fine-tune it on merely one training sample.

\oursnew{} also leverages an instance segmentation algorithm that provides a semantic prior
to guide the appearance model computed on the first frame.
This adds robustness to appearance changes of the object and in practice helps in keeping the quality throughout a longer period of the video.

The appearance model is combined with the semantic prior by means of a new conditional classifier
as a trainable module in a CNN.

\ifCLASSOPTIONcompsoc
  \section*{Acknowledgments}
\else
  \section*{Acknowledgment}
\fi
Research funded by the EU Framework Programme for Research and Innovation Horizon 2020 (Grant No. 645331, EurEyeCase), and by the Swiss Commission for Technology and Innovation (CTI, Grant No. 19015.1 PFES-ES, NeGeVA). The authors gratefully acknowledge support by armasuisse and thank NVidia Corporation for donating the GPUs used in this project.

\ifCLASSOPTIONcaptionsoff
  \newpage
\fi

\bibliographystyle{ieee}
\bibliography{main}

\begin{IEEEbiography}[{\vspace{-3mm}\includegraphics[width=1in,height=1.25in,clip,keepaspectratio]{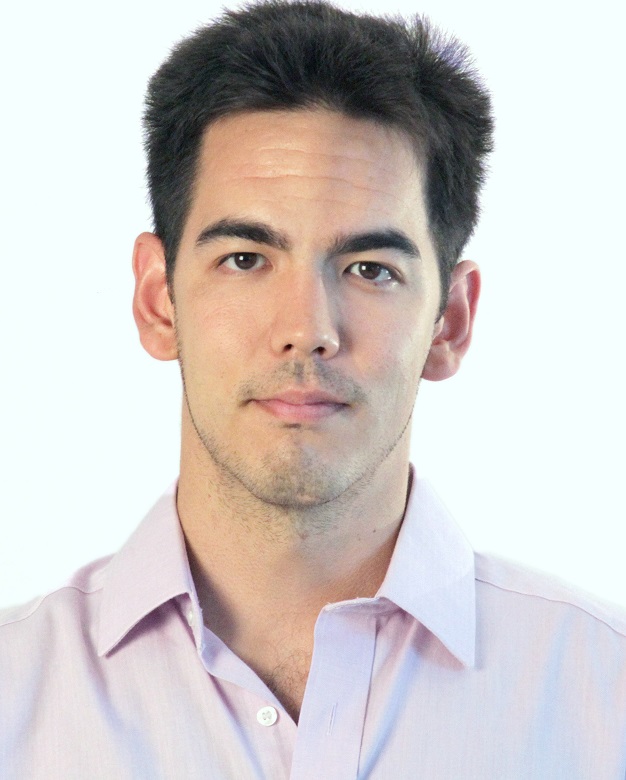}}]
{Kevis-Kokitsi Maninis} is a PhD candidate at ETHZ, Switzerland,
in Prof. Luc Van Gool's Computer Vision Lab (2015).
He received the Diploma degree in Electrical and Computer Engineering from National Technical University of Athens (NTUA) in 2014.
He worked as undergraduate research assistant in the Signal Processing and Computer Vision group of NTUA (2013-2014).
\end{IEEEbiography}

\begin{IEEEbiography}[{\vspace{-3mm}\includegraphics[width=1in,height=1.25in,clip,keepaspectratio]{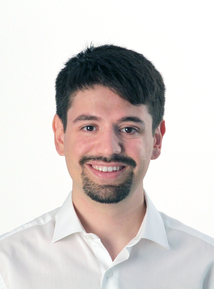}}]
{Sergi Caelles} is a Ph.D. candidate at ETHZ, Switzerland, in Prof. Luc Van Gool's Computer Vision Lab (2016). He received the degree in Electrical Engineering and the M.Sc. in Telecommunications Engineering from the Universitat Polit\`ecnica de Catalunya, BarcelonaTech (UPC). He worked at Bell Laboratories, New Jersey (USA) in 2014. His research interest include computer vision with special focus on video object segmentation and deep learning. 
\end{IEEEbiography}

\begin{IEEEbiography}[{\vspace{-3mm}\includegraphics[width=1in,height=1.25in,clip,keepaspectratio]{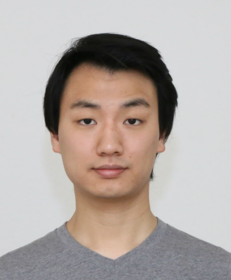}}]
{Yuhua Chen} is a PhD candidate at ETHZ, Switzerland, in Prof. Luc Van Gool's Computer Vision Lab (2015). He received a B.Sc in Physics from the University of Science and Technology of China (USTC) in 2013, and M.Sc in Electrical Engineering and Information Technology from ETH Z\"urich in 2015. His research interests lie in deep learning for semantic segmentation and object detection. 

\end{IEEEbiography}

\begin{IEEEbiography}[{\includegraphics[width=1in,height=1.25in,clip,keepaspectratio]{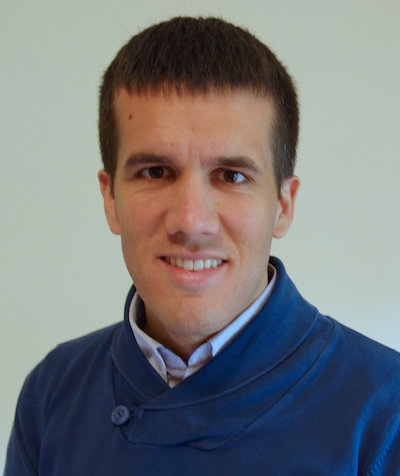}}]
{Jordi Pont-Tuset} is a post-doctoral researcher at ETHZ, Switzerland,
in Prof. Luc Van Gool's Computer Vision Lab (2015).
He received the degree in Mathematics in 2008, the degree in Electrical Engineering 
in 2008, the M.Sc. in Research on Information and Communication Technologies in 2010, and the Ph.D with honors in
2014; all from the Universitat Polit\`{e}cnica de Catalunya, BarcelonaTech (UPC).
He worked at Disney Research, Z\"urich (2014).
\end{IEEEbiography}

\begin{IEEEbiography}[{\includegraphics[width=1in,height=1.25in,clip,keepaspectratio]{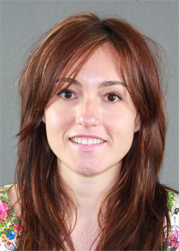}}]
{Laura Leal-Taix{\'e}} is leading the Dynamic Vision and Learning group at the Technical University of Munich, Germany. 
She received her Bachelor and Master degrees in Telecommunications Engineering from the Technical University of Catalonia (UPC), Barcelona. She did her Master Thesis at Northeastern University, Boston, USA and received her PhD degree (Dr.-Ing.) from the Leibniz University Hannover, Germany. 
During her PhD she did a one-year visit at the Vision Lab at the University of Michigan, USA. 
She also spent two years as a postdoc at the
Institute of Geodesy and Photogrammetry of ETH Zurich, Switzerland and one year at the Technical University of Munich.
Her research interests are dynamic scene understanding, in particular multiple object tracking and segmentation, as well as machine learning for video analysis.
\end{IEEEbiography}

\begin{IEEEbiography}[{\includegraphics[width=1in,height=1.25in,clip,keepaspectratio]{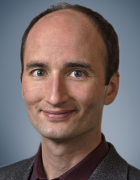}}]
{Daniel Cremers} received Bachelor degrees in Mathematics (1994) and Physics (1994), and a Master's degree in Theoretical Physics (1997) from the University of Heidelberg. In 2002 he obtained a PhD in Computer Science from the University of Mannheim, Germany. Since 2009 he holds the chair for Computer Vision and Pattern Recognition at the Technical University, Munich. His publications received several awards and he has obtained numerous and prestigious funding grants. He has served as area chair (associate editor) for ICCV, ECCV, CVPR, ACCV, IROS, etc, and as program chair for ACCV 2014. He serves as general chair for the European Conference on Computer Vision 2018 in Munich. On March 1st 2016, Prof. Cremers received the Leibniz Award 2016, the biggest award in German academia.
\end{IEEEbiography}

\begin{IEEEbiography}[{\includegraphics[width=1in,height=1.25in,clip,keepaspectratio]{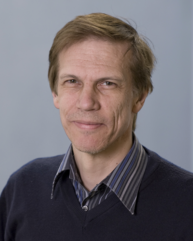}}]
{Luc Van Gool} got a degree in electromechanical engineering at the Katholieke Universiteit Leuven in 1981. Currently, he is professor at the Katholieke Universiteit Leuven, Belgium, and the ETHZ, Switzerland. He leads computer vision research at both places. His main interests include 3D reconstruction and modeling, object recognition, and tracking, and currently especially their confluence in the creation of autonomous cars. On the latter subject, he leads a large-scale project funded by Toyota. He has authored over 300 papers in this field. He has been a program chair or general chair of several major computer vision conferences.  He received several Best Paper awards, incl. a David Marr prize. In 2015, he received the 5-yearly excellence prize of the Flemish Fund for Scientific Research and, in 2016, a Koenderink Award. In 2017  he was nominated one of the main Tech Pioneers in Belgium by business journal `De Tijd', and one of the 100 Digital Shapers of 2017 by Digitalswitzerland. He is a co-founder of 10 spin-off companies.
\end{IEEEbiography}

\end{document}